\documentclass[12pt]{article}
\usepackage[T1]{fontenc}
\usepackage[utf8]{inputenc}
\usepackage{lmodern}
\usepackage{amsmath,amssymb,amsfonts,amsthm,mathtools}
\usepackage{booktabs,multirow,multicol,array}
\usepackage{graphicx,xcolor,float,placeins,microtype}
\usepackage{parskip}
\usepackage{etoolbox}
\usepackage[round]{natbib}
\usepackage[hyphens]{url}
\usepackage[hidelinks,unicode]{hyperref}
\IfFileExists{xurl.sty}{\usepackage{xurl}}{}
\def\spacingset#1{\renewcommand{\baselinestretch}{#1}\small\normalsize}
\AtBeginEnvironment{table}{\spacingset{1}}
\AtBeginEnvironment{table*}{\spacingset{1}}
\newcommand{\bs}{\boldsymbol}
\DeclareMathOperator{\DC}{DC}
\DeclareMathOperator{\dcov}{dcov}

\DeclareMathOperator{\softmax}{Softmax}
\newcommand{\DCdn}[1][n]{\widehat{\DC}_{#1}}
\newcommand{\indep}{\perp\!\!\!\perp}
\newcommand{\Rrep}{\bs R}

\theoremstyle{plain}
\newtheorem{theorem}{Theorem}[section]
\newtheorem{proposition}[theorem]{Proposition}

\newtheorem{corollary}[theorem]{Corollary}
\theoremstyle{definition}
\newtheorem{definition}[theorem]{Definition}
\newtheorem{assumption}[theorem]{Assumption}
\theoremstyle{remark}
\newtheorem{remark}[theorem]{Remark}

\title{\Large\bfseries Deep Fair Learning: Task-Aware Fair Representations via Joint Distance-Covariance Regularization}
\author{Enze Shi\textsuperscript{1}, Yiqun Xiao\textsuperscript{1}, Linglong Kong\textsuperscript{1},
Bei Jiang\textsuperscript{1}\thanks{Corresponding author. Email: \url{bei1@ualberta.ca}}}
\date{}
\hypersetup{pdftitle={Deep Fair Learning: Task-Aware Fair Representations via Joint Distance-Covariance Regularization},
pdfauthor={Enze Shi; Yiqun Xiao; Linglong Kong; Bei Jiang},
pdfkeywords={Group fairness; Representation learning; Joint estimation; Statistical calibration; Downstream retraining}}
\begin{document}
\spacingset{1}
\maketitle
\begin{center}
\textsuperscript{1}Department of Mathematical and Statistical Sciences, University of Alberta
\end{center}

\begin{abstract}
Ensuring fairness is essential as machine learning increasingly informs consequential decisions. However, many fairness-aware methods focus on the outputs of individual predictors, without directly controlling sensitive information retained in the underlying representations. We propose Deep Fair Learning (DFL), which combines distance covariance regularization with predictive loss to jointly learn representations and downstream predictors, promoting fairness at both levels while preserving task-relevant information. Its marginal and class-conditional formulations target independence and separation, respectively. Under suitable regularity conditions, we establish non-asymptotic joint excess-risk rates and convergence of the learned representation up to natural invariances. We further derive fairness-inheritance bounds linking representation-level dependence to downstream disparities over suitable predictor classes, extending fairness guarantees beyond the jointly trained predictor. Experiments on tabular, text, and image benchmarks show that DFL achieves lower fairness gaps than competing methods in many evaluated settings while maintaining competitive predictive accuracy, with fairness gains largely preserved after downstream retraining.

\end{abstract}


\spacingset{1.8}

\section{Introduction}
\label{sec:intro}

Machine learning increasingly informs consequential decisions in lending, hiring, and healthcare,
making fairness a central requirement alongside predictive accuracy~\citep{hardt2016equality,potash2015predictive,barocas2023fairness}.
Many fairness-aware methods constrain the outputs of an individual predictor through penalties,
adversaries, or post-processing~\citep{wan2023processing,caton2024fairness}. Such interventions can reduce
disparities in the fitted model while leaving sensitive information in the features it uses. When those
features are reused and the downstream predictor is retrained, that information remains available to
influence predictions. This motivates a broader objective: learn representations that support accurate
prediction while carrying fairness benefits beyond the jointly trained predictor.

Fair representation learning addresses this objective by controlling sensitive information in the feature
space~\citep{zemel2013learning,madras2018learning}. The central challenge is to reduce sensitive dependence
without discarding task-relevant information. Optimizing these goals separately can shape features that
are weakly associated with the sensitive attribute but poorly suited to the downstream task. Joint
learning provides a direct way to balance them: the predictive loss determines which information the
representation must preserve, while a representation-level penalty controls sensitive dependence.

We propose \emph{Deep Fair Learning} (DFL), a framework that jointly learns a nonlinear representation
map and a downstream predictor using predictive loss and distance covariance regularization. Marginal
and class-conditional penalties target independence and separation, respectively. Distance covariance
detects nonlinear dependence, accommodates discrete or continuous sensitive attributes, and admits a
minibatch estimator without an adversarial optimization or a kernel-bandwidth parameter. In our
fine-tuning implementation, a compact trainable encoder operates on frozen backbone features, making the
resulting representation available for downstream retraining.

Existing fair representation methods provide important foundations: adversarial approaches connect
representation alignment to downstream fairness~\citep{madras2018learning,zhao2020conditional}, while
kernel and distribution-matching methods characterize trade-offs and transfer
guarantees~\citep{tan2020learning,sadeghi2022tradeoff,oneto2020exploiting}. Distance-based approaches
also learn fair representations through variational objectives~\citep{liu2022fairdisco}.
A central statistical question is how jointly learning
a nonlinear encoder and its task predictor affects the estimation of both objects, and how the
resulting representation-level control governs predictors fitted afterwards.

DFL connects training, representation recovery, and downstream reuse through three guarantees.
A non-asymptotic bound controls the joint excess risk of the learned encoder and head at a fixed
fairness--utility trade-off. Under identifiability and local growth, this bound yields convergence
of the representation up to rigid transformations and calibration of predictive risk. Finally,
representation-level distance covariance bounds the group disparity of every replacement head
in a specified function class; the class-conditional version controls opportunity gaps.
To our knowledge, this is the first analysis of jointly trained, distance-covariance-regularized
neural representations to establish this combination of guarantees. Experiments across tabular,
text, and image data evaluate the same framework and its persistence after head replacement.
Table~\ref{tab:positioning} summarizes the distinctions; detailed related work appears in
Section~\ref{sec:related}.

\paragraph{Contributions.}
\begin{enumerate}
\item We develop a unified framework for
independence and separation that couples representation-level dependence control with downstream
predictive loss, preserving task-relevant information during fairness-aware training.
\item We establish non-asymptotic excess-risk rates for the jointly estimated encoder and head.
Under identifiability and local growth, the rates imply representation recovery up to rigid
transformations and convergence of predictive performance to the trade-off oracle.
\item We give uniform fairness-inheritance bounds for downstream heads with controlled
complexity. Marginal and class-conditional versions convert representation-level dependence
into quantitative score and opportunity guarantees after retraining.
\item Against strong prediction- and representation-level baselines, including RLACE, CFair-EO and
LAFTR-EO-CE, DFL ranks first or second in 30 of 48 displayed accuracy and fairness comparisons.
Downstream retraining largely preserves its gains, and controlled ablations examine the
representation-level penalty.
\end{enumerate}

\begin{table}[!htbp]
\centering
\caption{Comparison with related fairness methods. ``Joint'' denotes encoder--head training driven by
the task loss; ``Sep.'' denotes support for class-conditional fairness. Guarantees describe the published
formulations. The experiments evaluate the subset listed in Section~\ref{sec:protocol}.}
\label{tab:positioning}
\fontsize{9}{11}\selectfont
\setlength{\tabcolsep}{3pt}
\begin{tabular}{@{}>{\raggedright\arraybackslash}p{.20\linewidth}>{\raggedright\arraybackslash}p{.21\linewidth}>{\raggedright\arraybackslash}p{.15\linewidth}cc>{\raggedright\arraybackslash}p{.27\linewidth}@{}}
\toprule
\textbf{Method} & \textbf{Dependence measure} & \textbf{Level} & \textbf{Joint} & \textbf{Sep.} & \textbf{Guarantee} \\
\midrule
AdvDebias \citep{zhang2018mitigating}   & adversarial (implicit)      & output & --- & \checkmark & --- \\
FairMixup \citep{chuang2021fair}        & path-wise DP surrogate      & output & --- & \checkmark & --- \\
DRAlign \citep{li2023fairer}            & distributional alignment    & output & --- & \checkmark & --- \\
DiffMCDP \citep{jin2024on}              & smoothed MCDP surrogate     & output & --- & ---        & --- \\
\midrule
INLP \citep{ravfogel2020null}           & linear predictability       & repr.\ (post-hoc) & --- & --- & --- \\
RLACE \citep{ravfogel2022linear}        & linear adversarial game     & repr.\ (post-hoc) & --- & --- & linear-erasure \\
LEACE \citep{belrose2023leace}          & linear predictability       & repr.\ (post-hoc) & --- & --- & exact linear erasure \\
SUP \citep{shi2024debiasing}            & SDR subspace                & repr.\ (post-hoc) & --- & --- & subspace \\
\midrule
LAFTR \citep{madras2018learning} & adversarial alignment & repr. & \checkmark & \checkmark & downstream fairness \\
FairDisCo \citep{liu2022fairdisco} & distance-based dependence & repr.\ (VAE) & --- & --- & dependence analysis \\
FARE \citep{jovanovic2023fare} & group distributions & repr.\ (restricted) & --- & --- & fairness certificates \\
VFAE \citep{louizos2016vfae}            & MMD (Gaussian kernel)       & repr. & \checkmark & --- & --- \\
CFair \citep{zhao2020conditional}       & adversarial cond.\ alignment & repr. & \checkmark & \checkmark & trade-off lower bound \\
\citet{grari2020renyi}                  & HGR (adversarial)           & repr. & \checkmark & --- & --- \\
\citet{quadrianto2019discovering}       & HSIC (Gaussian kernel)      & data domain & \checkmark & \checkmark & --- \\
\citet{tan2020learning}                 & RKHS cross-covariance       & repr.\ (RKHS) & --- & \checkmark & exact for in-subspace heads \\
\citet{sadeghi2022tradeoff}             & kernel dependence           & repr.\ (RKHS) & --- & --- & closed-form optimum \\
\midrule
\textbf{DFL (this paper)}               & (conditional) DC            & repr. & \checkmark & \checkmark
& \shortstack[l]{joint risk; repr.\ convergence\\fairness inheritance} \\
\bottomrule
\end{tabular}
\end{table}

\FloatBarrier
\section{Related work}
\label{sec:related}

\paragraph{Output-level fairness.}
In-processing methods add fairness constraints or penalties to predictor training~\citep{zafar2019fairness,berk2023fair},
targeting demographic parity~\citep{dwork2012fairness,kamishima2012fairness}, equality of
opportunity~\citep{hardt2016equality}, or related criteria. Approaches include adversarial
training~\citep{zhang2018mitigating,han2021diverse}, interpolation~\citep{chuang2021fair}, distributional
alignment~\citep{li2023fairer}, and differentiable disparity surrogates~\citep{jin2024on}. These methods
control the predictions of the fitted model. DFL instead regularizes the representation consumed by the
predictor, enabling analysis of fairness after that predictor is replaced.

\paragraph{Fair representations and concept erasure.}
Fair representation learning includes early distribution-matching formulations~\citep{zemel2013learning},
adversarial censoring~\citep{edwards2016censoring,madras2018learning,elazar2018adversarial}, variational
approaches with maximum mean discrepancy~\citep{louizos2016vfae}, and information-theoretic
control~\citep{gupta2021controllable}. Post-hoc concept erasure modifies existing features: INLP iteratively
removes linearly predictive directions~\citep{ravfogel2020null}, RLACE uses a linear adversarial
formulation~\citep{ravfogel2022linear}, LEACE provides closed-form linear erasure~\citep{belrose2023leace},
and SUP removes a sufficient-dimension-reduction subspace~\citep{shi2024debiasing}. Kernel extensions
address nonlinear structure~\citep{ravfogel2022adversarial}. DFL learns a nonlinear encoder jointly with
the task predictor, allowing the predictive loss to guide which features are retained.

\paragraph{Dependence-minimizing encoders.}
The closest approaches directly reduce dependence between representations and sensitive attributes.
CFair uses class-conditional adversarial alignment and balanced error to target equalized odds and
accuracy parity~\citep{zhao2020conditional}; \citet{grari2020renyi} minimize adversarially estimated
Hirschfeld--Gebelein--R\'enyi maximal correlation. Kernel approaches offer complementary constructions.
\citet{tan2020learning} identify an RKHS subspace whose cross-covariance with sensitive attributes
vanishes, providing fairness for predictors within that subspace. \citet{sadeghi2022tradeoff} optimize an
RKHS-linear encoder in closed form using dependence statistics for utility and invariance, while
\citet{dehdashtian2024ufate} estimate achievable trade-offs to audit representations.
\citet{quadrianto2019discovering} use HSIC to learn fair translations in the original data domain.
Distance covariance is related to HSIC through distance-induced kernels~\citep{sejdinovic2013equivalence}.
DFL uses a directly computed distance statistic together with predictive loss, and analyzes estimation
of the nonlinear encoder and quantitative fairness inheritance for subsequently fitted heads.
The feasibility of fair representations is also subject to fundamental accuracy--fairness
limits~\citep{zhao2022inherent,mcnamara2019costs}.

\paragraph{Statistical guarantees and downstream reuse.}
LAFTR combines theoretical downstream-fairness guarantees with experiments on new
predictors and tasks~\citep{madras2018learning}. Multitask MMD/Sinkhorn methods derive transfer
bounds~\citep{oneto2020exploiting}, and FARE gives finite-sample fairness certificates for
restricted encoders~\citep{jovanovic2023fare}. These results establish important precedents for
theoretical and empirical analysis of fair representations. DFL's novelty concerns the combined
analysis of a jointly estimated nonlinear encoder and task head, representation convergence modulo
objective invariances, and quantitative inheritance under its distance-covariance penalty.

\section{Fair representation learning with DFL}
\label{sec:setup}\label{sec:method}

Let $(\bs X,Y,\bs Z)$ denote the input features, target, and sensitive attributes, with
$\bs X\in\mathbb R^p$, $Y\in\{1,\ldots,K\}$, and $\bs Z\in\mathbb R^d$.
The features may be supplied by a frozen backbone, and the sensitive attributes may be discrete,
continuous, or multivariate. DFL learns an encoder $g_\theta:\mathbb R^p\to\mathbb R^p$ and a head
$f_\phi:\mathbb R^p\to\mathbb R^K$. The representation is $\Rrep=g_\theta(\bs X)$;
the head produces logits, with class probabilities $\softmax(f_\phi(\Rrep))$ and prediction
$\widehat Y=\arg\max_k f_{\phi,k}(\Rrep)$. Write $\mathbf e_Y$ for the one-hot label vector.

\subsection{From prediction fairness to representation fairness}
\label{sec:rep-level}

Two widely used group-fairness criteria are \emph{independence}, $\widehat Y\indep\bs Z$, and
\emph{separation}, $\widehat Y\indep\bs Z\mid Y$~\citep{barocas2023fairness}. In binary classification
with a binary sensitive attribute $Z$, demographic parity (DP) requires
\begin{align}
\mathbb P(\widehat Y=1\mid Z=0)&=\mathbb P(\widehat Y=1\mid Z=1),\label{eq:dp-main}
\end{align}
while equalized odds (EO) requires, for each $y\in\{0,1\}$,
\begin{align}
\mathbb P(\widehat Y=1\mid Y=y,Z=0)&=\mathbb P(\widehat Y=1\mid Y=y,Z=1).\label{eq:eo-main}
\end{align}
DP equalizes positive-prediction probabilities across groups, whereas EO equalizes both true- and
false-positive rates. These are population fairness criteria. Our experiments quantify disparities
using the classwise TPR-RMS gap and the MCDP score-distribution metric, whose definitions and
multiclass aggregation are given in Appendix~\ref{appendix:metric}.

DFL applies these criteria to the representation itself:
\begin{equation}
\Rrep\indep\bs Z\quad\text{or}\quad\Rrep\indep\bs Z\mid Y.
\label{eq:rep-targets}
\end{equation}
This changes what the fairness intervention controls. If group-conditional representation distributions
agree, any predictor using only those representations sees the same distribution across groups.
Consequently, exact representation independence passes to every measurable downstream predictor;
the conditional version gives the same conclusion within each target class. Reducing dependence
therefore provides a route to fairness that can persist when the head is replaced. In finite samples
dependence is generally nonzero, so Section~\ref{sec:theory-inherit} quantifies the remaining disparity
for suitable predictor classes.

The two targets reflect different choices about which group differences to remove. Marginal
independence aligns representations across groups without conditioning on the outcome; separation
aligns them within each outcome class. When outcome frequencies differ between groups, these are
different statistical requirements. DFL makes this choice through the penalty rather than through
a change in architecture, so the same joint-learning framework can address either criterion.

\subsection{Measuring dependence with distance covariance}
\label{sec:dc}\label{app:estimators}\label{sec:depend}

Distance covariance detects general dependence between random vectors of possibly different
dimensions~\citep{szekely2007measuring,szekely2009brownian}. For vectors $\bs U,\bs V$ with finite
second moments, we use $\DC$ to denote the \emph{squared} distance covariance:
\begin{align}
\DC(\bs U,\bs V)
&=\mathbb E\big[\|\bs U-\bs U'\|\,\|\bs V-\bs V'\|\big]
-2\mathbb E\big[\|\bs U-\bs U'\|\,\|\bs V-\bs V''\|\big]\nonumber\\
&\quad+\mathbb E\|\bs U-\bs U'\|\;\mathbb E\|\bs V-\bs V'\|,
\label{eq:DC}
\end{align}
where primes indicate independent copies of the pair and $\|\cdot\|$ is the Euclidean norm.
It is nonnegative and equals zero precisely under independence. We write
$\dcov(\bs U,\bs V)=\sqrt{\DC(\bs U,\bs V)}$ when its square root is needed.

For separation, let $\rho_k=\mathbb P(Y=k)>0$ be the population class probabilities and
$\pi_k=\rho_k^4/\sum_j\rho_j^4$ the corresponding population penalty weights. Define
$\DC(\bs Z,\Rrep\mid Y)=\sum_k\pi_k\DC(\bs Z,\Rrep\mid Y=k)$.
Because every $\pi_k$ is positive, this quantity vanishes exactly when representation separation holds.
Definition \eqref{eq:DC} is the moment form of the squared distance covariance. It is
equivalent to the characteristic-function form that makes the characterization of independence transparent:
with $\psi_U,\psi_V,\psi_{U,V}$ the characteristic functions of $\bs U\in\mathbb R^{d}$,
$\bs V\in\mathbb R^{q}$ and the pair,
\begin{equation}
\DC(\bs U,\bs V)=\int_{\mathbb R^{d+q}}
\frac{\big|\psi_{U,V}(\bs t,\bs s)-\psi_U(\bs t)\psi_V(\bs s)\big|^2}{c_dc_q\|\bs t\|^{d+1}\|\bs s\|^{q+1}}\,
d\bs t\,d\bs s,
\qquad c_k=\frac{\pi^{(k+1)/2}}{\Gamma((k+1)/2)},
\label{eq:DC-cf}
\end{equation}
from which $\DC(\bs U,\bs V)\ge0$ with equality if and only if $\psi_{U,V}=\psi_U\psi_V$ almost everywhere,
i.e.\ if and only if $\bs U\indep\bs V$ \citep{szekely2007measuring}.

Given $n$ i.i.d.\ copies $\{(Z_i,R_i)\}_{i=1}^n$ of $(\bs Z,\Rrep)$, the unbiased $U$-statistic estimator of
the squared distance covariance \eqref{eq:DC} is
\begin{align}\label{eq:U-stat}
\DCdn(\bs Z,\Rrep)=\binom{n}{4}^{-1}\sum_{1\le i_1<i_2<i_3<i_4\le n}h\big((Z_{i_1},R_{i_1}),\dots,(Z_{i_4},R_{i_4})\big),
\end{align}
with the symmetric kernel
\begin{align*}
h\big((Z_1,R_1),\dots,(Z_4,R_4)\big)
&=\frac14\sum_{\substack{1\le i,j\le 4\\ i\neq j}}\|Z_i-Z_j\|\,\|R_i-R_j\|
\\&\quad+\frac1{24}\Big(\sum_{\substack{1\le i,j\le 4\\ i\neq j}}\|Z_i-Z_j\|\Big)\Big(\sum_{\substack{1\le k,l\le 4\\ k\neq l}}\|R_k-R_l\|\Big)\\
&\quad-\frac14\sum_{i=1}^4\ \sum_{\substack{1\le j\le 4\\ j\neq i}}\|Z_i-Z_j\|\ \sum_{\substack{1\le j\le 4\\ j\neq i}}\|R_i-R_j\|,
\end{align*}
where $\|\cdot\|$ is the Euclidean norm. The class-conditional
estimator aggregates within-class estimates,
\begin{align}\label{eq:U-stat-cond}
\DCdn(\bs Z,\Rrep\mid\bs Y)=\sum_{k:n_k\ge4} w_{n,k}\,\DCdn[n_k](\bs Z,\Rrep),
\qquad w_{n,k}=\binom{n_k}{4}\Big/\sum_{j=1}^K\binom{n_j}{4},
\end{align}
computed on $S_k=\{i:Y_i=k\}$ with $n_k=|S_k|$, provided
$\sum_j\binom{n_j}{4}>0$.

\begin{theorem}[Consistency of the dependence estimators]
\label{thm:converge}
For a fixed representation $g$, suppose $(\bs Z_i,\Rrep_i,Y_i)$ are i.i.d.,
$\mathbb E(\|\Rrep\|^2+\|\bs Z\|^2)<\infty$, and each of finitely many classes has positive
probability. Then, as $n\to\infty$,
\[
\DCdn(\bs Z,\Rrep)\xrightarrow{\mathcal P}\DC(\bs Z,\Rrep),\qquad
\DCdn(\bs Z,\Rrep\mid Y)\xrightarrow{\mathcal P}\DC(\bs Z,\Rrep\mid Y).
\]
\end{theorem}

This result justifies the population targets of
the empirical penalties; the data-dependent learned encoder is analyzed in Section~\ref{sec:theory-risk}.

\paragraph{Why distance covariance?}
DC detects nonlinear dependence and accommodates both discrete and continuous sensitive variables.
Its estimator uses pairwise distances directly, requiring neither a density model nor an adversarial
inner optimization or a kernel-bandwidth choice. These properties make it convenient for minibatch
training. It is related to the Hilbert--Schmidt independence criterion (HSIC) through
distance-induced kernels~\citep{sejdinovic2013equivalence},
and the associated energy-distance identity also supplies the link to downstream fairness.

\subsection{Joint learning of representations and predictions}
\label{sec:normalization}\label{sec:utility}

A fair representation must also retain the information needed for the target task: a constant map
would remove sensitive dependence but lose predictive utility. DFL balances these goals by training
the encoder and head together. With $\Rrep=g_\theta(\bs X)$ and cross-entropy
$\ell(y,u)=-\log\softmax(u)_y$, the marginal objective is
\begin{equation}
\mathcal L(\theta,\phi)=\mathbb E[\ell(Y,f_\phi(\Rrep))]+\lambda\DC(\bs Z,\Rrep)-\mu\DC(\mathbf e_Y,\Rrep),
\label{eq:pop-obj}
\end{equation}
where $\lambda>0$ and $\mu\ge0$. Replacing the sensitive-dependence term by its class-conditional
version gives the separation objective. The predictive loss shapes features for the actual downstream
task; the label-dependence term provides an additional informativeness incentive. The tuning
parameters determine the fairness--utility trade-off.

Joint optimization makes the predictive task part of representation learning itself. Gradients from
the head identify features useful for classification, while the dependence penalty discourages
sensitive structure in those features. The predictive loss directly measures how effectively the
head uses the retained label information, complementing the informativeness penalty.
Once the encoder is frozen, a replacement head uses features shaped by both requirements,
even though that new head need not include a fairness penalty in its own training objective.

\paragraph{Empirical objectives.}
The empirical objective under the independence criterion is
\begin{align}\label{eq:obj-ind}
\mathcal L_n(\theta,\phi)
&=\frac1n\sum_{i=1}^n\ell\big(Y_i,f_\phi(g_\theta(X_i))\big)
+\lambda\,\DCdn\big(\bs Z,g_\theta(\bs X)\big)
-\mu\,\DCdn\big(\mathbf e_{\bs Y},g_\theta(\bs X)\big),
\end{align}
and under the separation criterion
\begin{align}\label{eq:obj-sep}
\mathcal L_n(\theta,\phi)
&=\frac1n\sum_{i=1}^n\ell\big(Y_i,f_\phi(g_\theta(X_i))\big)
+\lambda\,\DCdn\big(\bs Z,g_\theta(\bs X)\mid\bs Y\big)
-\mu\,\DCdn\big(\mathbf e_{\bs Y},g_\theta(\bs X)\big).
\end{align}
Both are minimized jointly over $(\theta,\phi)$ by stochastic gradient descent. The reported experiments use
the separation form, in the single-parameter $\alpha$ form of \eqref{eq:as-run}.

The empirical objective $\mathcal L_n$ replaces the expectation by a sample average and DC by its
sample estimator. We optimize both networks with minibatch stochastic gradients. In the fine-tuning
implementation, a compact DenseNet-style encoder transforms frozen backbone features while preserving
their dimension. After training, the encoder can be retained and a new head fitted on its outputs.
Appendix~\ref{appendix:setting} gives the architecture, normalization, and training protocol.

The training and reuse stages have distinct roles. During training, every minibatch passes through
the encoder and head, and both networks receive gradients from the joint objective. For separation,
the fairness term compares sensitive dependence within target classes, while the predictive term uses
all observations in the batch. After training, predictions are computed from $g_\theta(\bs X)$
without requiring a separate sensitive-attribute input. Group annotations are used to construct
the training penalty and to evaluate disparities.
The trained encoder exports features in the original feature dimension, allowing a new classifier
for the same target to be trained independently. This separation of stages motivates the
head-replacement evaluation in Section~\ref{sec:rq2}.

\section{Theoretical analysis}
\label{sec:theory-risk}\label{app:proof-risk}

We study DFL at fixed fairness--utility weights and give a chain of guarantees for the fitted model:
a non-asymptotic joint-risk rate, representation and predictive calibration, and an end-to-end
fairness bound for subsequently fitted heads. The last guarantee includes the encoder's estimation
error as well as dependence at the population trade-off oracle.

\subsection{Joint estimation and excess risk}
\label{app:formal-risk}
Write $\mathcal L(g,f)$ for \eqref{eq:pop-obj} as a function of the encoder and head. Let $(g_0,f_0)$
minimize this objective over bounded measurable pairs, and let $(\widehat g_n,\widehat f_n)$ minimize
the empirical objective over bounded ReLU network classes $\mathcal G_n\times\mathcal F_n$.
The pair $(g_0,f_0)$ is the \emph{trade-off oracle}: the population optimum for the chosen balance
of predictive utility and fairness. Define the joint excess risk
\[
\Delta_n=\mathcal L(\widehat g_n,\widehat f_n)-\mathcal L(g_0,f_0).
\]
Boundedness controls the scale of the dependence penalties.

Here $n$ is the training sample size, $p$ is the feature and representation dimension, and $K$ is the
number of target classes. The dimensions $p$ and $K$ are fixed as $n$ grows.
The classes $\mathcal G_n$ and $\mathcal F_n$ contain the candidate encoders
and heads, respectively, with network size allowed to grow with $n$. A network class is specified by
its depth $D$, maximum width $W$, parameter count $S$, and output bound $B$. We write $B_g$ and $B_f$
for the encoder and head bounds. Coordinatewise clipping at $[-10,10]$ gives the experimental
encoder the Euclidean bound $B_g=10\sqrt p$, fixing a finite scale for dependence and inheritance.

Population losses average over a fresh observation from the data distribution. Because the empirical
minimizer depends on the training sample, $\Delta_n$ is random; the expectation in the excess-risk
bound also averages over that sample. The weights $(\lambda,\mu)$ stay fixed in this analysis, so
the comparison is always with the same population trade-off.

Let $\sigma(t)=\max\{t,0\}$ and let $\mathcal{NN}^{(r\to s)}_{D,W,S,B}$ denote ReLU feedforward networks
$\mathbb R^r\to\mathbb R^s$ of depth $D$, width $\le W$, at most $S$ parameters, and sup-norm at most $B$.
Set $\mathcal G_n=\mathcal{NN}^{(p\to p)}_{D_g,W_g,S_g,B_g}$ and
$\mathcal F_n=\mathcal{NN}^{(p\to K)}_{D_f,W_f,S_f,B_f}$, where the output norm is Euclidean and
$\|g\|_\infty=\mathrm{ess\,sup}_x\|g(x)\|_2$. Write $\mathcal L_n$ for the empirical version of $\mathcal L$. The joint empirical minimizer is:
\begin{equation}
(\widehat g_n,\widehat f_n)\in
\arg\min_{(g,f)\in\mathcal G_n\times\mathcal F_n}\mathcal L_n(g,f).
\label{eq:joint_erm}
\end{equation}
We also write $(g_0,f_0)=(g_{\lambda,\mu},f_{\lambda,\mu})$ to make the fixed trade-off weights explicit.
Throughout, $\mathcal L_{\rm DC}(g)=\lambda\DC(\bs Z,g(\bs X))-\mu\DC(\mathbf e_Y,g(\bs X))$,
so $\mathcal L=\mathcal L_{\rm ce}+\mathcal L_{\rm DC}$.

\begin{definition}[H\"older class]
\label{def:holder}
Let $\beta=k+a$ with $k\in\mathbb N_0$ and $a\in(0,1]$. The H\"older class $H^{\beta}(\mathcal D,B)$ consists
of functions $h:\mathcal D\to\mathbb R$ whose partial derivatives $\partial^\gamma h$ exist for all
$\|\gamma\|_1\le k$ and satisfy
\[
\max_{\|\gamma\|_1\le k}\|\partial^\gamma h\|_\infty\le B,
\qquad
\max_{\|\gamma\|_1=k}\ \sup_{x\neq x'}\frac{|\partial^\gamma h(x)-\partial^\gamma h(x')|}{\|x-x'\|_2^{a}}\le B .
\]
We write $L_{f_0}$ for the H\"older constant implied by $f_0\in H^{\beta_f}(\cdot,B_{f0})$.
\end{definition}

Standard asymptotic notation is used throughout: $a\wedge b=\min\{a,b\}$, $a\vee b=\max\{a,b\}$;
$a_n=\mathcal O(b_n)$, $a_n\lesssim b_n$ and $b_n\gtrsim a_n$ all mean
$\limsup_n|a_n/b_n|<\infty$; and $a_n\asymp b_n$ means both directions hold. For a multi-index
$\alpha\in\mathbb N_0^d$, $\partial^\alpha h$ denotes the corresponding partial derivative, and
$\|f\|_\infty$ the sup-norm.

\begin{assumption}[Bounded data]
\label{asm:data_bound}
$\mathrm{supp}(\bs X)\subseteq[0,1]^p$, $\|\bs Z\|\le B_Z$ and $\|\mathbf e_{\bs Y}\|\le B_Y$ almost surely.
\end{assumption}

\begin{assumption}[Oracle smoothness]
\label{asm:oracle_holder}
Each coordinate of $g_0$ lies in the H\"older class $H^{\beta_g}([0,1]^p,B_{g0})$ of
Definition~\ref{def:holder} and each coordinate of
$f_0$ lies in $H^{\beta_f}([-B_g,B_g]^p,B_{f0})$, for some $\beta_g,\beta_f>0$.
\end{assumption}

\begin{theorem}[Non-asymptotic joint excess risk]\label{thm:joint_excess_rate}\label{thm:main-risk}

For positive integers $M_g,N_g,M_f,N_f$, take
$D_g\asymp M_g\log(8M_g)$, $W_g\asymp N_g\log(8N_g)$,
$D_f\asymp M_f\log(8M_f)$, and $W_f\asymp N_f\log(8N_f)$.
Under Assumptions~\ref{asm:data_bound}--\ref{asm:oracle_holder}, for sufficiently large $n$,
\begin{align}
\mathbb E[\Delta_n]
&\le C_1\Big\{(M_fN_f)^{-2\beta_f/p}+(M_gN_g)^{-2s_f\beta_g/p}
+(M_gN_g)^{-2\beta_g/p}\Big\}\nonumber\\
&\quad+C_2 B_\ell\sqrt{\frac{(M_g+M_f)(N_g^2M_g+N_f^2M_f)
\log(2+N_g^2M_g+N_f^2M_f)\log^5 n}{n}}\nonumber\\
&\quad+C_2B_g(\lambda B_Z+\mu B_Y)
\sqrt{\frac{M_g^2N_g^2\log(2+N_g^2M_g)\log^5 n}{n}}.
\label{eq:joint_excess_general}
\end{align}
where $s_f=\min\{\beta_f,1\}$ and $B_\ell=\log K+2B_f$. The constants $C_1,C_2$ depend only on
$(\beta_g,\beta_f,p,K)$, the bounds $(B_g,B_f,B_Z,B_Y)$, the weights $(\lambda,\mu)$,
and the oracle regularity constants $(L_{f_0},B_{g0},B_{f0})$. 
Taking $M_g,M_f$ fixed and choosing $N_g\asymp n^{p/(4\beta_g+2p)}$ and
$N_f\asymp n^{p/(4\beta_f+2p)}$ gives the rate in
\eqref{eq:joint_excess_rate} up to polylogarithmic factors.

\begin{equation}
\mathbb E[\Delta_n]\lesssim
n^{-\beta_f/(2\beta_f+p)}+n^{-s_f\beta_g/(2\beta_g+p)}+n^{-\beta_g/(2\beta_g+p)},
\label{eq:joint_excess_rate}
\end{equation}
where the rate is understood up to polylogarithmic factors. In subsequent rate conversions, $r_n$
denotes the upper bound including those factors from \eqref{eq:joint_excess_general}.
\end{theorem}

The bound separates approximation of the oracle pair from sampling error in the predictive loss and
dependence estimates. The encoder's approximation affects both the penalty and the composed predictor,
which explains the two representation terms. When $\beta_f\ge1$, these terms have the same order,
leaving one rate for learning the encoder and one for the head.

The smoothness indices describe the regularity of the oracle functions: a H\"older order between
zero and one controls how quickly function values vary, while a larger order additionally controls
derivatives. The factor $s_f=\min\{\beta_f,1\}$ enters because perturbing the representation changes
the input to the head. Network size balances the ability to approximate these smooth functions
against the statistical complexity of estimating them. The constants hidden by $\lesssim$ depend
on the fixed smoothness, dimension, output bounds, and trade-off weights; the displayed powers
describe how the error decreases with sample size.

More data improves estimation of the population optimum selected by the weights. The next result
connects this objective-level convergence to recovery of the representation and its predictive
performance.

For an oracle head with at least Lipschitz regularity ($\beta_f\ge1$), the representation terms have the same rate.
The result describes the statistical performance of an empirical minimizer; it does not certify that
stochastic gradient descent reaches that minimizer.

\subsection{Representation convergence and downstream performance}
\label{subsec:quotient_calibration}\label{sec:downstream}\label{app:formal-calibration}

Small excess risk concerns an objective value, whereas representation convergence concerns a function.
The latter requires accounting for the objective's invariances: rotating, reflecting, or translating
a representation leaves its pairwise distances unchanged, and an inverse transformation of the head
preserves predictions. We therefore compare representations after the best rigid alignment.

Let $P_X$ denote the marginal distribution of $\bs X$, and let
$\mathbb H=L^2(P_X;\mathbb R^p)$, with
$\|g\|_{\mathbb H}=(\mathbb E\|g(\bs X)\|_2^2)^{1/2}$, and let
$\mathbb G=\mathbb O(p)\ltimes\mathbb R^p$ be the rigid-motion group, acting by
$((O,c)\cdot g)(x)=Og(x)+c$.

\begin{definition}[Representation equivalence and distance]
\label{def:quotient_distance}
Write $g_1\sim g_2$ when $g_2=Og_1+c$ almost surely for some $(O,c)\in\mathbb G$, and denote the
equivalence class by $[g]$. Its quotient distance is
\begin{equation}
d_\sim([g_1],[g_2])=\inf_{(O,c)\in\mathbb G}\|g_1-(O,c)\cdot g_2\|_{\mathbb H}.
\label{eq:quotient_dist_main}
\end{equation}
\end{definition}

Profiling out the head gives
\begin{equation}
\mathcal L^\star(g):=\inf_{\|f\|_\infty\le B_f}\mathcal L(g,f),\qquad
\mathcal L^\star_{\min}:=\inf_{\|g\|_\infty\le B_g}\mathcal L^\star(g).
\label{eq:profiled_L_main}
\end{equation}
The profiled objective is constant on rigid-motion orbits. We impose identifiability of its minimizing
orbit and a local growth condition to convert excess risk into representation error.

\begin{assumption}[Tradeoff-oracle uniqueness and orbit identifiability]
\label{asm:quotient_regularity}
Given $(\lambda,\mu)$, there exists a joint minimizer $(g_0,f_0)$ of $\mathcal{L}$ over bounded measurable functions
$\|g\|_\infty\le B_g$ and $\|f\|_\infty\le B_f$, such that for any other joint minimizer $(g,f)$ under the same bounds,
we have $g\sim g_0$.
Moreover, for every $\varepsilon>0$,
\begin{equation}
\delta(\varepsilon)
:=
\inf\Big\{\mathcal{L}^\star(g)-\mathcal{L}^\star_{\min}:\ \|g\|_\infty\le B_g,\ d_{\sim}([g],[g_0])\ge \varepsilon\Big\}
\;>\;0.
\label{eq:orbit_separation_main}
\end{equation}
\end{assumption}

\begin{assumption}[Local power growth on the quotient]
\label{asm:quotient_local_growth}
There exist constants $c_{\rm LG}>0$, $\kappa>0$, and $r_{\rm LG}>0$ such that for all measurable $g$ with $\|g\|_\infty\le B_g$,
\begin{equation}
d_{\sim}\big([g],[g_0]\big)\le r_{\rm LG}
\quad\Longrightarrow\quad
\mathcal{L}^\star(g)-\mathcal{L}^\star_{\min}
\;\ge\;
c_{\rm LG}\,d_{\sim}\big([g],[g_0]\big)^{\kappa}.
\label{eq:local_growth_main}
\end{equation}
\end{assumption}

\begin{theorem}[Quotient-space calibration: consistency and rate conversion]
\label{thm:quotient_calibration_main}\label{thm:main-calibration}
Under  Assumption~\ref{asm:quotient_regularity}, let $(\widehat g_n,\widehat f_n)$ be any estimator with $\|\widehat f_n\|_\infty\le B_f$ and $\|\widehat g_n\|_\infty\le B_g$, and define
\[
\Delta_n:=\mathcal{L}(\widehat g_n,\widehat f_n)-\mathcal{L}(g_0,f_0)
=\mathcal{L}(\widehat g_n,\widehat f_n)-\mathcal{L}^\star_{\min}.
\]
(i) (\emph{Consistency}) If $\Delta_n\to 0$ in probability, then
\[
d_{\sim}\big([\widehat g_n],[g_0]\big)\to 0
\qquad\text{in probability.}
\]
In particular, if $\mathbb{E}[\Delta_n]\to 0$, then $d_{\sim}([\widehat g_n],[g_0])\to 0$ in probability.

(ii) (\emph{Rates}) If additionally Assumption~\ref{asm:quotient_local_growth} holds and $\Delta_n=O_{\mathbb{P}}(r_n)$ for some $r_n\downarrow 0$, then
\[
d_{\sim}\big([\widehat g_n],[g_0]\big)=O_{\mathbb{P}}\big(r_n^{1/\kappa}\big).
\]
\end{theorem}

By Markov's inequality, \eqref{eq:joint_excess_rate} and the local growth condition give
$d_\sim([\widehat g_n],[g_0])=O_{\mathbb P}((r_n)^{1/\kappa})$
whenever $r_n\to0$.

The exponent $\kappa$ describes the local shape of the profiled objective around the oracle orbit.
For quadratic growth, $\kappa=2$, an objective error of order $r_n$ yields representation error of
order $\sqrt{r_n}$. The notation $O_{\mathbb P}(r_n)$ means that the error divided by $r_n$ is bounded
in probability. Here $r_n$ can be chosen as the excess-risk bound from Theorem~\ref{thm:main-risk},
including its logarithmic factors. This conversion gives a direct interpretation of how improved
joint estimation sharpens representation recovery.

\subsection{Downstream classification performance}
\label{app:formal-utility}

Let $\mathcal L_{\rm ce}(g,f)=\mathbb E[\ell(Y,f(g(\bs X)))]$. Representation convergence separates
predictive utility from the joint objective.

\begin{theorem}[Downstream cross-entropy calibration to the tradeoff oracle]
\label{thm:ce_calibration_main}
Work under Assumption~\ref{asm:quotient_regularity} and Assumption~\ref{asm:data_bound}.
Let $\Delta_n$ be as in Theorem~\ref{thm:quotient_calibration_main}, then
\begin{equation}
\big|\mathcal{L}_{\rm ce}(\widehat g_n,\widehat f_n)-\mathcal{L}_{\rm ce}(g_0,f_0)\big|
\;\le\;
\Delta_n
\;+\;
16\big(\lambda B_Z+\mu B_Y\big)\,d_{\sim}\big([\widehat g_n],[g_0]\big).
\label{eq:ce_calibration_main}
\end{equation}

Consequently, if $\Delta_n\to 0$ in probability, then
$\big|\mathcal{L}_{\rm ce}(\widehat g_n,\widehat f_n)-\mathcal{L}_{\rm ce}(g_0,f_0)\big|\to 0$ in probability.
If moreover Assumption~\ref{asm:quotient_local_growth} holds and $\Delta_n=O_{\mathbb{P}}(r_n)$ for some $r_n\downarrow 0$, then
\[
\big|\mathcal{L}_{\rm ce}(\widehat g_n,\widehat f_n)-\mathcal{L}_{\rm ce}(g_0,f_0)\big|
=
O_{\mathbb{P}}(r_n)+O_{\mathbb{P}}(r_n^{1/\kappa}).
\]
\end{theorem}

The comparison is with the optimum at the chosen fairness--utility trade-off. A standard log-loss
calibration also gives a Bayes classification bound with separate trade-off bias and estimation terms
(Corollary~\ref{cor:bayes_01_bias_main}).

In \eqref{eq:ce_calibration_main}, $B_Z$ and $B_Y$ bound the norms of the sensitive vector and one-hot label. In particular,
$B_Y=1$, and a sensitive vector with $d$ binary coordinates admits $B_Z=\sqrt d$.
These constants make the dependence on the attribute encoding explicit. The first term is
joint estimation error; the second accounts for the difference between the learned and oracle
representations. This explains why representation convergence is useful beyond identifying the
encoder: it separates predictive performance from an objective that also contains fairness and
informativeness penalties. Both errors are measured relative to the same trade-off oracle, making
their contributions to predictive calibration explicit.

\paragraph{Classification risk and trade-off bias.}
\label{app:proof-utility-bound}
To convert cross-entropy control into a Bayes classification bound, define
\[
p_n(x):=\operatorname{softmax}\!\big(\widehat f_n(\widehat g_n(x))\big)\in\Delta^K,
\qquad
\mathcal{L}_{\log}(p):=\mathbb{E}\big[-\log p_Y(\bs X)\big],
\]
so that $\mathcal{L}_{\log}(p_n)=\mathcal{L}_{\rm ce}(\widehat g_n,\widehat f_n)$.
Let $\eta(x):=(\mathbb{P}(Y=1\mid \bs X=x),\dots,\mathbb{P}(Y=K\mid \bs X=x))$ and
$\mathcal{L}_{\log}^\star:=\mathbb{E}[-\log \eta_Y(\bs X)]$,
and let $\mathcal{R}_{0\text{-}1}^\star$ be the Bayes $0$--$1$ risk.

\begin{corollary}[Bayes $0$--$1$ excess risk and irreducible tradeoff bias]
\label{cor:bayes_01_bias_main}
Let $\widehat h_n(x):=\arg\max_k p_{n,k}(x)$ and $\mathcal{R}_{0\text{-}1}(\widehat g_n,\widehat f_n):=\mathbb{P}(\widehat h_n(\bs X)\neq Y)$.
Then
\begin{align}
\mathcal{R}_{0\text{-}1}(\widehat g_n,\widehat f_n)-\mathcal{R}_{0\text{-}1}^\star
&\;\le\;
\sqrt{2\Big(\mathcal{L}_{\rm ce}(\widehat g_n,\widehat f_n)-\mathcal{L}_{\log}^\star\Big)}\nonumber\\
&\;\le\;
\sqrt{2\big(\mathcal{L}_{\rm ce}(g_0,f_0)-\mathcal{L}_{\log}^\star\big)}
+
\sqrt{2|\mathcal{L}_{\rm ce}(\widehat g_n,\widehat f_n)-\mathcal{L}_{\rm ce}(g_0,f_0)|}.
\label{eq:bayes_01_bias_main}
\end{align}
\end{corollary}

\begin{remark}[Interpretation of the bias term]
The first term in \eqref{eq:bayes_01_bias_main} is an irreducible bias induced by the fixed fairness--utility tradeoff $(\lambda,\mu)$:
even the tradeoff oracle $(g_0,f_0)$ need not achieve the Bayes log-loss $\mathcal{L}_{\log}^\star$.
The second term is an estimation term, controlled by Theorem~\ref{thm:ce_calibration_main}, and it vanishes as the joint excess risk vanishes.
This decomposition makes explicit how fairness constraints can create unavoidable degradation in predictive accuracy relative to Bayes,
while still allowing consistent learning of the best predictor under the imposed tradeoff.
\end{remark}

\subsection{Fairness of the learned representation}
\label{app:formal-fairness}\label{sec:fairness-bound}\label{app:proof-fairness-bound}
\begin{theorem}[Fairness of the learned representation]
\label{thm:fairness-bound}
Let $\lambda>0$. Under Assumptions~\ref{asm:data_bound} and \ref{asm:quotient_regularity}, for any $(\widehat g_n,\widehat f_n)$ with
$\|\widehat g_n\|_\infty\le B_g$, $\|\widehat f_n\|_\infty\le B_f$,
\begin{equation}
\DC\big(\bs Z,\widehat g_n(\bs X)\big)
\;\le\;
\DC\big(\bs Z,g_0(\bs X)\big)
\;+\;\frac{2\Delta_n}{\lambda}
\;+\;\frac{16\big(\lambda B_Z+2\mu B_Y\big)}{\lambda}\,d_\sim\big([\widehat g_n],[g_0]\big).
\label{eq:fairness-bound}
\end{equation}
Under Assumptions~\ref{asm:quotient_regularity}--\ref{asm:quotient_local_growth} and \eqref{eq:joint_excess_rate}, the last two terms are
$O_{\mathbb P}\big((r_n)/\lambda\big)+O_{\mathbb P}\big((r_n)^{1/\kappa}\big)$.
The separation criterion obeys the same bound with $\DC(\bs Z,\cdot)$ replaced by
$\DC(\bs Z,\cdot\mid\bs Y)$ and a class-weighting constant in place of $B_Z$; see
Proposition~\ref{prop:fairness-bound-sep} and the remark following it.
\end{theorem}

At fixed $\lambda$, the bound approaches the oracle's dependence level; increasing sample size
controls estimation error without requiring that level to be zero. Section~\ref{sec:theory-inherit}
translates this representation guarantee into fairness bounds for replacement heads.

\begin{assumption}[Nonvanishing cells]
\label{asm:cells}
$\bs Y$ takes finitely many values with $\rho_k=\mathbb P(\bs Y=k)>0$ for all $k$, and
$p_k:=\mathbb P(\bs Z=1\mid\bs Y=k)\in(0,1)$ for all $k$; write $q_k=1-p_k$ and
$c_\star=\min_k\,p_kq_k>0$.
\end{assumption}

For the separation objective, $(g_0,f_0)$ and $\Delta_n$ refer to its own trade-off oracle and excess risk.

\begin{proposition}[Separation version]
\label{prop:fairness-bound-sep}
Let $\lambda>0$, let Assumptions~\ref{asm:data_bound} and \ref{asm:cells} hold and let $\mathcal L_{\rm DC}$ be formed with
the conditional penalty $\DC(\bs Z,g(\bs X)\mid\bs Y)=\sum_k\pi_k\DC(\bs Z,g(\bs X)\mid\bs Y=k)$. Then
\[
\DC\big(\bs Z,\widehat g_n(\bs X)\mid\bs Y\big)
\le\DC\big(\bs Z,g_0(\bs X)\mid\bs Y\big)
+\frac{2\Delta_n}{\lambda}
+\frac{16\big(\lambda C_w B_Z+2\mu B_Y\big)}{\lambda}\,d_\sim\big([\widehat g_n],[g_0]\big),
\]
where $C_w:=\sum_{k=1}^K\pi_k/\mathbb P(\bs Y=k)$.
\end{proposition}

\begin{remark}[Reading the constant $C_w$]
For the class-proportional weights $\pi_k\propto\mathbb P(\bs Y=k)^4$ of \eqref{eq:U-stat-cond},
$C_w=\sum_k\mathbb P(\bs Y=k)^3/\sum_j\mathbb P(\bs Y=j)^4$; for balanced weights $\pi_k=1/K$,
$C_w=K^{-1}\sum_k\mathbb P(\bs Y=k)^{-1}$. Both equal $K$ when the classes are balanced, and $C_w\ge1$
always. They differ in how they respond to skew: the balanced weighting blows up when any class is rare
(the $\mathbb P(\bs Y=k)^{-1}$ term), whereas the class-proportional weighting shrinks toward $1$ under
extreme skew because it puts almost all of the penalty mass on the dominant class --- which is precisely why
it gives a weak guarantee for rare classes, as discussed after Theorem~\ref{thm:inheritance-sep}. Neither
weighting is uniformly preferable; the choice should follow which classes the separation guarantee is meant
to protect.
\end{remark}

\subsection{Fairness inheritance after downstream retraining}
\label{sec:theory-inherit}\label{app:proof-inheritance}\label{app:formal-inheritance}

Representation-level control matters because the trained head need not be the final consumer of the
features. Exact independence, marginal or conditional, passes to any measurable head. For approximate
independence, the head's complexity matters: a sufficiently sensitive decision rule can amplify a small
remaining difference between group distributions. The following bound quantifies that dependence.

Let $\|\Rrep\|\le B_g$ almost surely. For the distance kernel
$\kappa(u,v)=\tfrac12(\|u\|+\|v\|-\|u-v\|)$ on $\mathcal B_g=\{r:\|r\|\le B_g\}$,
let $\mathbb H_\kappa$ denote its reproducing-kernel Hilbert space (RKHS).
This space provides a predictor class with explicit disparity control, and is also denoted by
$\mathbb H_{\rm dist}$. In this subsection, $p$ denotes group probability.

\begin{theorem}[Fairness inheritance under head replacement: independence]
\label{thm:inheritance}\label{thm:main-inheritance}
Let $\bs Z\in\{0,1\}$ with $p=\mathbb P(\bs Z=1)\in(0,1)$, $q=1-p$. For a head $h:\mathcal B_g\to\mathbb R$
write $\Gamma(h)=\big|\mathbb E[h(\Rrep)\mid\bs Z=1]-\mathbb E[h(\Rrep)\mid\bs Z=0]\big|$. Then for every
$h\in\mathbb H_\kappa$ with $\|h\|_{\mathbb H_\kappa}\le M$,
\begin{equation}
\Gamma(h)\;\le\;\frac{M}{2pq}\,\sqrt{\DC\big(\bs Z,\Rrep\big)}\;=\;\frac{M}{2pq}\,\dcov\big(\bs Z,\Rrep\big).
\label{eq:inheritance}
\end{equation}
\end{theorem}

This guarantee applies simultaneously to all heads in the RKHS ball, including heads fitted after the
representation is frozen. Predictor complexity and group imbalance enter through $M/(2pq)$. Hard
decisions additionally require approximation by the stated RKHS class;
Theorem~\ref{thm:inheritance-sep} quantifies the within-class approximation remainder.

\paragraph{Class-conditional fairness.}
\begin{theorem}[Fairness inheritance: separation]
\label{thm:inheritance-sep}
Under Assumption~\ref{asm:cells}, let $\DC_k:=\DC(\bs Z,\Rrep\mid\bs Y=k)$ and let
$\DC(\bs Z,\Rrep\mid\bs Y)=\sum_k\pi_k\DC_k$ be the weighted conditional distance covariance of
Section~\ref{sec:dc} with weights $\pi_k>0$. Then for every $h\in\mathbb H_\kappa$ with
$\|h\|_{\mathbb H_\kappa}\le M$ and every class $k$,
\begin{align*}
&\big|\mathbb E[h(\Rrep)\mid\bs Z=1,\bs Y=k]-\mathbb E[h(\Rrep)\mid\bs Z=0,\bs Y=k]\big|\\
&\qquad\le\frac{M}{2p_kq_k}\sqrt{\DC_k}
\le\frac{M}{2c_\star}\sqrt{\frac{\DC(\bs Z,\Rrep\mid\bs Y)}{\pi_k}} .
\end{align*}
For binary $\bs Y$ and a head $h$ with values in $[0,1]$, read as the probability of the positive decision
given $\Rrep$, the left-hand side at $k=1$ is $\Delta_{\rm TPR}(h)$ and at $k=0$ is $\Delta_{\rm FPR}(h)$, and hence
$\Delta_{\rm EO}(h)\le\frac{M}{2c_\star}\sqrt{\DC(\bs Z,\Rrep\mid\bs Y)/\min_k\pi_k}$. For a hard decision
rule $D$, approximation by the norm-bounded RKHS class under the conditional law given $\bs Y=k$ adds
$2A_k(M,D)$ to the bound for class $k$, where
$A_k(M,D):=\inf_{\|h\|_{\mathbb H_\kappa}\le M}\max_{z}\mathbb E[|D(\Rrep)-h(\Rrep)|\mid\bs Z=z,\bs Y=k]$.
\end{theorem}

The factor $\pi_k^{-1/2}$ makes the cost of underweighting rare classes explicit.
The displayed constants apply to binary sensitive groups.

\paragraph{From estimation to downstream fairness.}
Combining the representation bound from Section~\ref{sec:fairness-bound} with marginal inheritance
gives a guarantee for the estimated representation and any replacement head in the specified class.

\begin{corollary}[End-to-end guarantee for retrained heads]
\label{cor:endtoend}
Under the assumptions of Theorem~\ref{thm:fairness-bound}, let $\bs Z\in\{0,1\}$ with $p=\mathbb P(\bs Z=1)\in(0,1)$, $q=1-p$. For every $h\in\mathbb H_\kappa$ with
$\|h\|_{\mathbb H_\kappa}\le M$, fitted on $\widehat g_n(\bs X)$ by any procedure,
\[
\Gamma(h)\;\le\;\frac{M}{2pq}\left\{\DC\big(\bs Z,g_0(\bs X)\big)+\frac{2\Delta_n}{\lambda}
+\frac{16(\lambda B_Z+2\mu B_Y)}{\lambda}\,d_\sim\big([\widehat g_n],[g_0]\big)\right\}^{1/2}.
\]
For the separation objective, Proposition~\ref{prop:fairness-bound-sep} and
Theorem~\ref{thm:inheritance-sep} give the within-class analogue. Probability-valued heads yield
randomized-decision TPR and FPR bounds; deterministic decisions additionally require approximation
by the norm-bounded RKHS class, as in Theorem~\ref{thm:inheritance-sep}.
\end{corollary}

Thus, if the oracle has zero sensitive dependence and the joint excess risk is
$O_{\mathbb P}(r_n)$, the gap for any fixed norm budget $M$ is
$O_{\mathbb P}((r_n+r_n^{1/\kappa})^{1/2})$.

\section{Experiments}
\label{sec:experiments}

The principal contribution is the joint-estimation and fairness-inheritance analysis. The
experiments assess competitiveness with representative fairness methods across data modalities,
persistence after replacing the trained head, and the effect of the objective's components.
Appendix~\ref{appendix:numeric} reports additional ablations, sensitivity studies, and multi-attribute results.

\subsection{Datasets and evaluation protocol}
\label{sec:protocol}

We use two tabular datasets, Adult (income prediction, sensitive attribute gender) and Bank (deposit
subscription, sensitive attribute age); two text datasets, BIOS (occupation prediction, gender) and
MOJI (sentiment prediction, dialect group); and CelebA (facial-attribute prediction with
gender, hair-color, or skin-color annotations). BIOS uses BERT features, MOJI uses DeepMoji
features~\citep{felbo2017deepmoji}, and CelebA uses extracted image features. Feature
dimensions and method-specific architectures are detailed in Appendices~\ref{app:datasets}
and~\ref{app:comparison-details}.

The main benchmark tables use DFL's separation objective. Encoder outputs are clipped
coordinatewise to $[-10,10]$ before entering both the dependence penalties and downstream head.
Predictive and informativeness terms receive weight $\alpha$; conditional fairness receives
$1-\alpha$. All methods use a common validation rule: minimize average TPR-RMS and MCDP
subject to at least $95\%$ of displayed STD accuracy, or select the most accurate candidate
if none qualifies. DFL searches $\alpha\in\{0.1,\ldots,0.9\}$; details are in
Appendix~\ref{appendix:setting}.

We compare with an unconstrained model (STD), four prediction-level methods (AdvDebias, FairMixup,
DRAlign, DiffMCDP), three representation-erasure methods (INLP, RLACE, SUP), and the conditional
adversarial representation methods CFair-EO and LAFTR-EO-CE. Results report
accuracy, classwise TPR-RMS (labelled TPR), and MCDP, all on a percentage scale. Zero TPR-RMS
is equivalent to equalized odds for binary targets; its multiclass form summarizes diagonal
class-conditional gaps. MCDP assesses marginal score disparity related to demographic parity.
Parentheses give standard deviations; bold and underline mark the best and second-best
fairness-aware means. Metric and baseline details are in Appendices~\ref{appendix:metric}
and~\ref{app:comparison-details}.

\subsection{Fairness--utility comparison}
\label{sec:rq1}

Tables~\ref{tab:adult-bank}--\ref{tab:celebA-male} compare fairness and predictive accuracy across
the three data modalities; Appendix~\ref{appendix:numeric} completes the CelebA pairs.
Across these 16 settings, DFL ranks first or second among fairness-aware methods in
30 of 48 metric comparisons: 11 accuracy, 10 TPR-gap, and 9 MCDP-gap comparisons.
On Adult, DFL reduces MCDP from $12.46$ to $7.31$ and has accuracy $79.24$ versus $77.67$
for the logistic-regression STD reference. On Bank, accuracy remains close to STD
($90.55$ versus $90.69$), and MCDP
falls from $11.31$ to $8.23$. DFL has the highest accuracy among fairness-aware methods on both
tabular datasets; on Bank its MCDP gap is within $0.01$ point of the smallest reported gap.

The text benchmarks show a stronger effect on both fairness metrics. On BIOS, DFL reduces the TPR
gap from $15.59$ to $9.01$, with accuracy $77.94$ compared with STD's $79.79$. On MOJI, it reduces
the TPR gap from $38.95$ to $10.47$ and MCDP from $39.86$ to $8.41$, while accuracy rises from
$71.19$ to $74.59$. DFL attains the smallest TPR and MCDP gaps on MOJI, while CFair-EO has the
highest accuracy there. On BIOS, CFair-EO has the smallest gaps on both metrics and DFL ranks second.

Table~\ref{tab:celebA-male} reports the image tasks with gender as the sensitive attribute. DFL has
the lowest MCDP gap on all four tasks and the lowest TPR gap for Attractive. For Wavy Hair, the TPR gap
falls from $29.14$ to $3.49$ and MCDP from $48.13$ to $18.04$, with accuracy changing from $83.52$
to $80.10$. On Smiling, DiffMCDP has the lowest TPR gap, while DFL has the lowest MCDP gap.
Additional sensitive attributes are evaluated in Appendix~\ref{appendix:numeric}.

\begin{table}[t]
\centering
\caption{Predictive accuracy and fairness on Adult and Bank. Parentheses give standard deviations. Among fairness-aware methods, best in \textbf{bold}, second best
\underline{underlined}.}
\label{tab:adult-bank}
\fontsize{7.8}{8.8}\selectfont
\begin{minipage}{0.47\textwidth}
\centering \textbf{Adult}\vspace{0.1cm}
\begin{tabular}{l@{\hskip 3pt}r@{\hskip 3pt}r@{\hskip 3pt}r}
\toprule
\textbf{Method} & \textbf{Accuracy $\uparrow$} & \textbf{TPR $\downarrow$} & \textbf{MCDP $\downarrow$} \\
\midrule
STD       & 77.67 & 3.55 & 12.46 \\
AdvDebias & 76.49(0.65) & \underline{3.27(0.28)} & 29.32(2.31) \\
FairMixup & 74.48(0.43) & \textbf{3.14(0.23)} & 24.91(1.58) \\
DRAlign   & 75.01(0.41) & 3.66(0.17) & 22.04(1.22) \\
DiffMCDP  & 73.93(0.31) & 3.28(0.38) & 11.50(1.09) \\
INLP      & 68.27(0.57) & 3.96(0.44) & 8.72(1.31) \\
RLACE     & 72.79(0.83) & 3.82(0.16) & 7.45(0.86) \\
SUP       & 70.53(0.36) & 3.52(0.28) & 8.16(1.25) \\
CFair-EO  & \underline{77.69(0.98)} & 3.49(1.00) & \textbf{6.60(1.26)} \\
LAFTR-EO-CE & 77.29(0.71) & 3.36(1.62) & \underline{6.62(0.30)} \\
\midrule
\textbf{DFL} & \textbf{79.24(0.33)} & 3.43(0.23) & 7.31(0.43) \\
\bottomrule
\end{tabular}
\end{minipage}
\hspace{1em}
\begin{minipage}{0.47\textwidth}
\centering \textbf{Bank}\vspace{0.1cm}
\begin{tabular}{l@{\hskip 3pt}r@{\hskip 3pt}r@{\hskip 3pt}r}
\toprule
\textbf{Method} & \textbf{Accuracy $\uparrow$} & \textbf{TPR $\downarrow$} & \textbf{MCDP $\downarrow$} \\
\midrule
STD       & 90.69 & 2.42 & 11.31 \\
AdvDebias & 60.56(1.00) & 2.20(0.13) & 17.55(1.39) \\
FairMixup & 59.48(1.96) & 2.21(0.24) & 13.18(2.94) \\
DRAlign   & 59.12(1.56) & 1.96(0.21) & 13.61(1.23) \\
DiffMCDP  & 60.01(1.83) & 2.18(0.13) & 11.00(0.97) \\
INLP      & 70.13(0.86) & 2.32(0.35) & 10.25(0.68) \\
RLACE     & 72.51(0.53) & 2.21(0.26) & 8.83(0.35) \\
SUP       & 70.82(0.47) & 2.46(0.21) & 9.69(0.42) \\
CFair-EO  & 88.65(0.01) & \textbf{1.66(0.07)} & \textbf{8.22(1.32)} \\
LAFTR-EO-CE & \underline{88.82(0.24)} & \underline{1.78(1.09)} & 10.23(2.34) \\
\midrule
\textbf{DFL} & \textbf{90.55(0.67)} & 2.25(0.16) & \underline{8.23(0.21)} \\
\bottomrule
\end{tabular}
\end{minipage}
\end{table}

\begin{table}[t]
\centering
\caption{Predictive accuracy and fairness on BIOS and MOJI. Parentheses give standard deviations. Bold and underline mark the best and second-best fairness-aware means.}
\label{tab:bios-moji}
\fontsize{7.8}{8.8}\selectfont
\begin{minipage}{0.47\textwidth}
\centering \textbf{BIOS}\vspace{0.1cm}
\begin{tabular}{l@{\hskip 3pt}r@{\hskip 3pt}r@{\hskip 3pt}r}
\toprule
\textbf{Method} & \textbf{Accuracy $\uparrow$} & \textbf{TPR $\downarrow$} & \textbf{MCDP $\downarrow$} \\
\midrule
STD       & 79.79 & 15.59 & 6.67 \\
AdvDebias & 77.51(1.32) & 13.51(1.24) & 10.89(1.33) \\
FairMixup & 75.01(0.83) & 12.58(1.43) & 11.72(1.29) \\
DRAlign   & 74.68(1.15) & 11.36(1.31) & 9.27(0.83) \\
DiffMCDP  & 75.18(1.30) & 10.04(1.03) & 6.42(0.76) \\
INLP      & 71.72(1.04) & 9.68(0.44) & 6.70(0.56) \\
RLACE     & 77.15(0.87) & 13.15(0.94) & 6.51(0.63) \\
SUP       & 76.36(0.59) & 12.64(0.33) & 6.63(0.41) \\
CFair-EO  & \underline{78.39(1.59)} & \textbf{8.87(1.13)} & \textbf{6.13(1.23)} \\
LAFTR-EO-CE & \textbf{78.82(0.73)} & 12.44(0.15) & 6.51(0.84) \\
\midrule
\textbf{DFL} & 77.94(0.34) & \underline{9.01(0.59)} & \underline{6.24(0.51)} \\
\bottomrule
\end{tabular}
\end{minipage}
\hspace{1em}
\begin{minipage}{0.47\textwidth}
\centering \textbf{MOJI}\vspace{0.1cm}
\begin{tabular}{l@{\hskip 3pt}r@{\hskip 3pt}r@{\hskip 3pt}r}
\toprule
\textbf{Method} & \textbf{Accuracy $\uparrow$} & \textbf{TPR $\downarrow$} & \textbf{MCDP $\downarrow$} \\
\midrule
STD       & 71.19 & 38.95 & 39.86 \\
AdvDebias & 70.95(1.14) & 30.41(1.46) & 28.77(2.43) \\
FairMixup & 69.71(1.21) & 26.95(2.39) & 23.85(2.69) \\
DRAlign   & 67.34(1.59) & 21.41(1.75) & 18.71(1.26) \\
DiffMCDP  & 68.55(0.83) & 15.97(0.73) & 14.14(0.91) \\
INLP      & 62.37(1.41) & 15.28(0.64) & 17.75(1.29) \\
RLACE     & 72.75(0.86) & 15.21(0.47) & 17.36(1.71) \\
SUP       & 69.25(0.78) & \underline{10.65(0.62)} & 18.35(1.43) \\
CFair-EO  & \textbf{75.02(0.36)} & 11.85(2.28) & \underline{9.99(2.90)} \\
LAFTR-EO-CE & 73.38(0.74) & 12.82(4.49) & 12.77(3.63) \\
\midrule
\textbf{DFL} & \underline{74.59(0.34)} & \textbf{10.47(0.53)} & \textbf{8.41(1.67)} \\
\bottomrule
\end{tabular}
\end{minipage}
\end{table}

\begin{table}[t]
\centering
\caption{Predictive accuracy and fairness on CelebA with $\bs Z=$ gender. Parentheses give
standard deviations. Best and second-best
fairness-aware means are bold and underlined. Other sensitive attributes are in Appendix~\ref{appendix:numeric}.}
\label{tab:celebA-male}
\fontsize{7.5}{8.5}\selectfont
\setlength{\tabcolsep}{2.5pt}
\begin{tabular}{lc|cccccc}
\toprule
$\bs Y$ & Metric & STD & DFL & DiffMCDP & RLACE & CFair-EO & LAFTR-EO-CE \\
\midrule
\multirow{3}{*}{Young}
& Accuracy $\uparrow$ & 85.01 & 79.48(0.92) & 79.15(0.74) & 81.91(1.22) & \textbf{82.10(0.81)} & \underline{82.08(0.43)} \\
& TPR Gap $\downarrow$ & 26.70 & \underline{2.43(2.01)} & 12.29(1.14) & 6.59(2.45) & 2.70(1.43) & \textbf{2.22(1.74)} \\
& MCDP Gap $\downarrow$ & 49.74 & \textbf{11.07(2.04)} & 20.79(2.38) & 26.47(2.78) & 17.73(5.54) & \underline{11.92(3.36)} \\
\midrule
\multirow{3}{*}{Attractive}
& Accuracy $\uparrow$ & 78.32 & \textbf{75.53(1.87)} & 74.79(0.85) & \underline{75.43(0.95)} & 74.59(0.48) & 74.67(0.49) \\
& TPR Gap $\downarrow$ & 40.50 & \textbf{2.89(1.31)} & 18.66(1.45) & 15.15(3.12) & \underline{7.84(2.34)} & 8.38(2.96) \\
& MCDP Gap $\downarrow$ & 57.20 & \textbf{21.15(2.59)} & \underline{23.91(2.67)} & 30.42(3.15) & 27.11(2.08) & 28.39(2.44) \\
\midrule
\multirow{3}{*}{Smiling}
& Accuracy $\uparrow$ & 80.51 & 77.76(1.67) & 77.81(0.68) & 77.53(0.87) & \textbf{79.58(0.53)} & \underline{78.92(0.81)} \\
& TPR Gap $\downarrow$ & 7.65 & 5.42(1.86) & \textbf{3.52(0.81)} & \underline{4.75(1.87)} & 5.07(0.64) & 5.13(0.91) \\
& MCDP Gap $\downarrow$ & 16.63 & \textbf{7.09(1.64)} & 9.95(1.45) & 8.84(1.96) & \underline{8.33(1.44)} & 8.92(2.77) \\
\midrule
\multirow{3}{*}{Wavy Hair}
& Accuracy $\uparrow$ & 83.52 & 80.10(1.08) & 79.76(0.79) & \underline{80.45(1.15)} & \textbf{81.27(0.67)} & 80.15(0.50) \\
& TPR Gap $\downarrow$ & 29.14 & \underline{3.49(2.38)} & 13.40(1.32) & 10.08(2.83) & 3.53(2.21) & \textbf{3.08(1.34)} \\
& MCDP Gap $\downarrow$ & 48.13 & \textbf{18.04(2.79)} & 21.12(2.42) & 25.62(2.69) & 19.80(4.48) & \underline{19.45(2.59)} \\
\bottomrule
\end{tabular}
\end{table}

\subsection{Persistence under downstream retraining}
\label{sec:rq2}

We freeze the learned representation, discard the jointly trained head, and fit a fresh one-hidden-layer
classifier using predictive loss alone. Table~\ref{tab:persistence} shows that accuracy changes by at most
$0.65$ percentage points across the five reported datasets. Most large disparity reductions persist:
the fresh head retains $98.4\%$ of the TPR-gap reduction on MOJI, $92.2\%$ on BIOS, and $92.2\%$ on
CelebA gender/Wavy Hair. MCDP reduction is also largely preserved on Adult, Bank, MOJI, and CelebA.

Across the tabular and text benchmarks, retraining changes the TPR gap by at most $0.51$ points and
MCDP by at most $0.71$ points. The corresponding changes on CelebA are $2.00$ and $1.50$ points,
relative to the substantial reductions from the original STD gaps. Table~\ref{tab:persistence}
reports each comparison. These results demonstrate empirical persistence after head replacement;
they do not evaluate the numerical bound of Theorem~\ref{thm:main-inheritance}, as explained in
Appendix~\ref{app:comparison-details}.

\begin{table}[t]
\centering
\caption{Persistence of fairness gains after downstream retraining. ``DFL head'' uses the jointly trained classifier
$f_{\hat\phi}$; ``fresh head'' discards it and trains an unconstrained one-hidden-layer classifier on the
frozen representation $g_{\hat\theta}$. $\delta$ is (fresh head $-$ DFL head). The CelebA row uses
$\bs Z=$ gender, $\bs Y=$ Wavy Hair; the remaining CelebA cells are in
Appendix~\ref{appendix:numeric}. Means over 20 replications.}
\label{tab:persistence}
\fontsize{8}{9.5}\selectfont
\setlength{\tabcolsep}{3.2pt}
\begin{tabular}{@{}l|ccc|ccc|ccc@{}}
\toprule
& \multicolumn{3}{c|}{\textbf{Accuracy} $\uparrow$}
& \multicolumn{3}{c|}{\textbf{TPR gap} $\downarrow$}
& \multicolumn{3}{c}{\textbf{MCDP gap} $\downarrow$} \\
\textbf{Dataset} & STD & DFL head & fresh head ($\delta$)
                 & STD & DFL head & fresh head ($\delta$)
                 & STD & DFL head & fresh head ($\delta$) \\
\midrule
Adult  & 77.67 & 79.24 & 79.32 ($+0.08$) & 3.55 & 3.43 & 3.47 ($+0.04$) & 12.46 & 7.31 & 7.52 ($+0.21$) \\
Bank   & 90.69 & 90.55 & 90.13 ($-0.42$) & 2.42 & 2.25 & 2.73 ($+0.48$) & 11.31 & 8.23 & 8.66 ($+0.43$) \\
BIOS   & 79.79 & 77.94 & 77.83 ($-0.11$) & 15.59 & 9.01 & 9.52 ($+0.51$) & 6.67 & 6.24 & 6.83 ($+0.59$) \\
MOJI   & 71.19 & 74.59 & 74.16 ($-0.43$) & 38.95 & 10.47 & 10.93 ($+0.46$) & 39.86 & 8.41 & 9.12 ($+0.71$) \\
CelebA & 83.52 & 80.10 & 79.45 ($-0.65$) & 29.14 & 3.49 & 5.49 ($+2.00$) & 48.13 & 18.04 & 19.54 ($+1.50$) \\
\bottomrule
\end{tabular}
\end{table}

\subsection{What drives the fairness--utility trade-off?}
\label{sec:rq3}\label{sec:rq4}\label{sec:rq5}

On BIOS, representation- and output-level DC penalties share the predictive loss. At the same
tuning, the representation penalty achieves $77.85\%$ accuracy versus $64.78\%$, with MCDP
gaps $6.52$ versus $7.76$; the corresponding TPR gaps are $8.97$ and $5.12$.
Appendix~\ref{sec:ablation-studies} reports the full ablation. Further sweeps vary the
label-informativeness term, trade-off weight, encoder capacity, and sensitive attributes
(Appendices~\ref{app:sensitivity} and~\ref{app:multi-celeba}). They show how DFL's
tunable components shape the observed fairness--utility trade-off.

\FloatBarrier
\section{Conclusion}
\label{sec:conclusion}

DFL jointly learns predictive representations and downstream classifiers while controlling sensitive
dependence at the representation level. Its marginal and conditional formulations address independence
and separation within one framework. We prove a non-asymptotic joint excess-risk rate; under
identifiability and local growth, it yields representation recovery up to rigid transformations and
predictive calibration to the trade-off oracle. A uniform inheritance bound controls the group
score disparity of suitable replacement heads by representation dependence. For the learned encoder,
the end-to-end guarantee adds oracle dependence and finite-sample estimation error, with a
class-conditional counterpart for opportunity gaps. The bound also makes the roles of
replacement-head complexity and group prevalence explicit. Across three data modalities, DFL ranks
first or second in 30 of 48 reported fairness-aware accuracy and gap comparisons, and substantial
gains persist after downstream retraining. Together, the guarantees and empirical results establish a foundation for
learning fair, predictive representations and reusing them with downstream classifiers.

\bibliography{bibliography}
\bibliographystyle{agsm}

\clearpage
\appendix
\numberwithin{equation}{section}
\numberwithin{figure}{section}
\numberwithin{table}{section}
\setcounter{equation}{0}\setcounter{figure}{0}\setcounter{table}{0}
\section*{Appendix}

\section{Evaluation metrics}\label{appendix:metric}
Let $\bs Y$ and $\widehat{\bs Y}$ be the one-hot true label and predicted class decision,
respectively, and let $q_j\in[0,1]$ be the predicted probability of class $j$.
The sensitive attribute is binary, $\bs Z\in\{0,1\}$. TPR-RMS uses class decisions;
MCDP uses probability scores.

The TPR gap for class $j$ is defined as:
\begin{align*}
    \text{TPR}_{z,j} = P(\hat{\bs{Y}}_j=1 \mid \bs{Z}=z, \bs{Y}_j=1), \quad \text{TPR}_{gap,j} = \text{TPR}_{1,j} - \text{TPR}_{0,j}.
\end{align*}
The TPR gap across $K$ classes is then computed as:
\begin{align*}
    \text{TPR}_{gap} = \sqrt{\frac{1}{K} \sum_{j=1}^K \text{TPR}_{gap,j}^2}\times100\%.
\end{align*}

The MCDP metric is defined as:
\begin{align*}
    \text{MCDP}_j = \max_{y \in [0,1]} \big| F_{1,j}(y) - F_{0,j}(y) \big|, \quad \text{MCDP}_{gap} = \sqrt{\frac{1}{K-1} \sum_{j=1}^{K-1} \text{MCDP}_{j}^2}\times100\%,
\end{align*}
where $F_{z,j}(y)=P(q_j\le y\mid\bs Z=z)$ is the group-conditional score CDF.
We follow the benchmark convention of averaging the first $K-1$ class-score gaps.
For binary targets, either probability column gives the same exact CDF supremum. Equality
of the two group score distributions implies demographic parity for every fixed score
threshold. For multiclass argmax decisions, matching each marginal score distribution
does not by itself match the joint score distribution or guarantee decision parity.
The tabular STD reference and CFair-EO/LAFTR-EO-CE evaluations use a $100$-bin
approximation on $[0,1]$.

\subsection{Separation metrics}
\label{app:sep-metrics}

The theory of Section~\ref{sec:theory-inherit} controls the true-positive-rate gap, the false-positive-rate gap
and their maximum. For binary $\bs Y$ and binary $\bs Z$ these are
\begin{align*}
\Delta_{\rm TPR}&=\big|\mathbb P(\widehat{\bs Y}{=}1\mid \bs Y{=}1,\bs Z{=}1)-\mathbb P(\widehat{\bs Y}{=}1\mid \bs Y{=}1,\bs Z{=}0)\big|,\\
\Delta_{\rm FPR}&=\big|\mathbb P(\widehat{\bs Y}{=}1\mid \bs Y{=}0,\bs Z{=}1)-\mathbb P(\widehat{\bs Y}{=}1\mid \bs Y{=}0,\bs Z{=}0)\big|,
\end{align*}
\[
\Delta_{\rm EO}=\max\{\Delta_{\rm TPR},\Delta_{\rm FPR}\},
\qquad
\Delta_{\rm EO}^{\rm sum}=\Delta_{\rm TPR}+\Delta_{\rm FPR}.
\]
Separation holds if and only if $\Delta_{\rm EO}=0$. For $K$-class problems, replacing
$\mathbb P(\widehat{\bs Y}_j{=}1\mid \bs Y_j{=}1,\bs Z{=}z)$ by its class-$j$ analogue and taking the RMS
across $j$, as in the TPR-RMS definition above, gives the multiclass versions.

For binary targets, the class-$0$ TPR gap is the negative of the class-$1$ FPR gap, so
$\mathrm{TPR\text{-}RMS}=\sqrt{(\Delta_{\rm TPR}^2+\Delta_{\rm FPR}^2)/2}\times100\%$.
It is zero if and only if equalized odds holds, although it is a different aggregation
from the maximum $\Delta_{\rm EO}$. For multiclass targets, the reported TPR-RMS
summarizes diagonal entries of the class-conditional confusion matrices and does not
assess every off-diagonal disparity. MCDP is a marginal score-distribution metric for
demographic parity, not a class-conditional separation metric.

\subsection{Multi-attribute reporting}
\label{app:multi-metrics}

For a vector-valued $\bs Z=(\bs Z^{(1)},\dots,\bs Z^{(m)})$ we report the per-coordinate gaps, that is,
$\Delta_{\rm TPR}$ and MCDP computed for each $\bs Z^{(j)}$ separately with the other coordinates marginalized
out. This is a marginal audit. The \emph{intersectional} quantities, which we do not report, are the
subgroup-conditional gaps
\[
\Delta_{\rm TPR}^{\rm sub}=\max_{s,s'\in\mathcal S}\big|\mathbb P(\widehat{\bs Y}{=}1\mid \bs Y{=}1,\bs Z{=}s)-\mathbb P(\widehat{\bs Y}{=}1\mid \bs Y{=}1,\bs Z{=}s')\big|,
\]
where $\mathcal S$ enumerates the $2^m$ attribute combinations, together with the sample size of each
subgroup. Equalizing every marginal gap does not bound $\Delta_{\rm TPR}^{\rm sub}$: a model can match rates on each
attribute separately while serving a small intersection badly. Accordingly, the multi-attribute experiment assesses marginal fairness rather than intersectional fairness. Methods for discovering such
subgroups are surveyed by \citet{xu2024intersectional}.

\section{Experimental settings}\label{appendix:setting}

\subsection{Datasets and frozen input features}
\label{app:datasets}

The five benchmarks cover tabular classification, text classification, and image attribute prediction.
The target $Y$ specifies the task, while $\bs Z$ specifies the groups used in the fairness penalty and
evaluation. Backbone features remain fixed during DFL training; the additional representation network
learns a task-aware transformation of those features. Table~\ref{tab:network-parameters} gives the
feature dimensions and encoder settings used in the reported experiments.

\paragraph{Adult.}
Adult is a census-income benchmark~\citep{kohavi1996scaling}. The binary target indicates whether
annual income exceeds \$50,000, and the sensitive attribute is the dataset's binary sex annotation,
referred to as gender in the result tables. The tabular feature representation has dimension $101$.
This task tests whether fairness regularization can retain income-predictive information while reducing
disparities between the annotated groups.

\paragraph{Bank.}
Bank contains records from a Portuguese bank's marketing campaigns~\citep{moro2014data}. The task is
to predict whether a client subscribes to a term deposit. The sensitive variable is the binary age group
$Z=\mathbf 1\{\mathrm{age}\ge25\}$, and the input feature dimension is $62$. The threshold defines
the groups used by both the conditional penalty and the reported fairness metrics.

\paragraph{BIOS.}
Bias in Bios contains biographies annotated with gender and one of $28$ occupations~\citep{de2019bias}.
We follow the dataset split of \citet{de2019bias}, predict the occupation, and use gender as the
sensitive attribute. Frozen BERT representations have $768$ coordinates. The multiclass setting
requires dependence control within occupation classes; accordingly, the reported TPR metric aggregates
class-specific group gaps using the RMS definition in Appendix~\ref{appendix:metric}.

\paragraph{MOJI.}
MOJI is a tweet-sentiment task with happy/sad labels from emoji annotations and African-American
English (AAE) versus Standard American English (SAE) group annotations~\citep{blodgett2016demographic}.
These dialect labels are a demographic proxy, rather than self-reported race. The training mixture is
AAE--happy $40\%$, SAE--happy $10\%$, AAE--sad $10\%$, and SAE--sad $40\%$~\citep{han2021diverse}, creating an association
between sentiment and dialect group. The reported experiments use frozen, $2{,}304$-dimensional
DeepMoji representations~\citep{felbo2017deepmoji}. Thus, BIOS and MOJI use different text encoders.

\paragraph{CelebA.}
CelebA contains $202{,}599$ celebrity face images with $40$ binary attribute annotations~\citep{liu2015deep}.
We consider four separate binary prediction tasks: Young, Attractive, Smiling, and Wavy Hair.
Each task is evaluated with Male, Black Hair, or Pale Skin as the sensitive annotation; these names
refer to the dataset's attributes. A frozen ResNet-18 supplies $512$-dimensional features. In the
multiple-attribute setting, the three sensitive annotations form a vector and the penalty acts on their
joint conditional dependence with the representation. Reported gaps remain per-attribute summaries,
as specified in Appendix~\ref{appendix:metric}.

\subsection{Encoder outputs and objective}
Let $\widetilde g_\theta(x)$ be the encoder output before clipping. The representation used by both
the head and the dependence penalties is
\[
g_{\theta,j}(x)=\min\{10,\max\{-10,\widetilde g_{\theta,j}(x)\}\},\qquad j=1,\ldots,p.
\]
Hence $\|g_\theta(x)\|_2\le10\sqrt p$. This supplies a Euclidean representation bound
$B_g=10\sqrt p$ for the boundedness conditions in the analysis; $10$ is the coordinate bound.
The theoretical restrictions on the oracle and head class are stated separately in
Section~\ref{app:formal-risk}.

The experimental objective is
\begin{equation}
\begin{aligned}
\mathcal L_n^{\alpha}(\theta,\phi)
&=\alpha\Big[\tfrac1n\textstyle\sum_i\ell\big(Y_i,f_\phi(g_\theta(X_i))\big)
-\DCdn\big(\mathbf e_{\bs Y},g_\theta(\bs X)\big)\Big]\\
&\quad +(1-\alpha)\,\DCdn\big(\bs Z,g_\theta(\bs X)\mid\bs Y\big)
\end{aligned}
\label{eq:as-run}
\end{equation}
for $\alpha\in(0,1)$. Dividing by $\alpha$ yields the separation objective with
$\lambda=(1-\alpha)/\alpha$ and $\mu=1$. All ablated objectives below use the same clipped
representation; only the included objective terms change.

\subsection{Network architecture}

We describe the structure of the DFL framework, which consists of a representation network \( g_{\theta}(\cdot) \) followed by a downstream classifier \( f_{\phi}(\cdot) \). The entire framework is built upon a modified DenseNet architecture. The key input parameters of the DFL framework are:

\begin{itemize}
    \item \textbf{Input Dimension} (\( p \)): The dimensionality of the input representation \( \bs{X} \).
    \item \textbf{Output Dimension} (\( K \)): The number of classes for the target variable \( \bs{Y} \).
    \item \textbf{Growth Rate}: Controls the number of features added by each layer in the dense blocks.
    \item \textbf{Depth}: Specifies the total number of layers in the network.
    \item \textbf{Reduction Factor}: Determines the compression applied in transition layers.
\end{itemize}

The network consists of three key components: input processing, dense blocks with transition layers, and output generation. The input is a \( p \)-dimensional vector, processed through a fully connected layer that maps it to an initial feature space of size \( 2 \times \text{growthRate} \). 

The network includes three dense blocks, each composed of multiple layers with dense connectivity. Each layer compresses features into an intermediate space of width \( 4 \times \text{growthRate} \), and its output is concatenated with the inputs of all preceding layers. After each dense block, a transition layer reduces the feature dimensions using batch normalization and a fully connected layer, controlled by the reduction factor.

Following the final dense block, a batch normalization layer and a fully connected layer produce the latent
representation of dimensionality \( p \). Each coordinate is then clipped to $[-10,10]$ as specified
above, and the clipped representation is passed through another fully connected layer to generate
logits for \( K \)-class classification. Softmax gives class probabilities, and log-softmax supplies
the log-probabilities used in the loss.

\begin{table*}[!htbp]
\centering
\caption{DFL benchmark configurations.}
\label{tab:network-parameters}
\fontsize{8.5}{9.5}\selectfont
\begin{tabular}{lcccccc}
\toprule
\textbf{Dataset} & \textbf{\( p \)} & \textbf{\( K \)} & \textbf{Growth Rate} & \textbf{Depth} & \textbf{Reduction Factor} & $\bs{\alpha}$ \\
\midrule
Adult   & 101 & 2  & 20 & 10 & 0.2 & 0.3 \\
Bank    & 62  & 2  & 20 & 10 & 0.2  & 0.1\\
BIOS    & 768 & 28 & 64 & 10 & 0.2 & 0.7\\
MOJI    & 2304 & 2  & 64 & 10 & 0.2 & 0.2\\
CelebA  & 512 & 2 & 64 & 10 & 0.2 & 0.5\\
\bottomrule
\end{tabular}
\end{table*}

\subsection{Hyperparameter selection}

For the reported DFL runs, training is set to 100 epochs with learning rate $10^{-3}$.
Figures~\ref{fig:tabular} and~\ref{fig:nlp} show the observed trajectories. Batch size is
$128$ except on BIOS, where it is $256$ to include more of its 28 classes in each batch
for the conditional DC terms. Other methods use the implementations described below.

The one-parameter objective is \eqref{eq:as-run}. We select $\alpha$ on validation data from
$\{0.1,0.2,\ldots,0.9\}$, minimizing the average TPR-RMS and MCDP gaps subject to accuracy at least
$95\%$ of the displayed STD validation accuracy. The same rule applies to each fairness-aware
method's candidate set; when no candidate meets the floor, the highest-validation-accuracy
candidate is selected. The five-point sweep in Table~\ref{tab:alpha} is a separate sensitivity
study, rather than the nine-point selection grid.
Network weights use Kaiming initialization and biases are initialized to zero.
Table~\ref{tab:network-parameters} lists the dataset-specific architecture and selected trade-off weight.

\subsection{Baseline implementations and comparison protocol}\label{app:comparison-details}

The Adult and Bank STD rows use the standard multinomial
logistic-regression configuration. Their reported MCDP uses $100$ histogram bins on $[0,1]$.
The CFair-EO and LAFTR-EO-CE rows train the same vector DenseNet encoder with
class-conditional balanced adversaries. They use a training-only three-fold weight search and
validation-selected checkpoints for Adult, Bank, BIOS, and MOJI. The CelebA rows use separate
$(\bs Y,\bs Z)$ pairs, one fixed adversarial weight, and a held-out validation checkpoint.

For baseline methods, we use the official code released by the respective authors. Specifically, for AdvDebias, FairMixup, DRAlign, and DiffMCDP, we use the implementation provided in \cite{jin2024on}. For INLP, we set the number of iterations to 100. For RLACE, we follow the training hyperparameters used in the original paper. For SUP, we adopt the same number of slices (\( p+1 \)) as in our method and select the dimension to remove by 10-fold cross-validation under the common validation rule above.

CFair-EO and LAFTR-EO-CE share a vector DenseNet encoder; the other baselines use their
method-specific settings. INLP, RLACE, and SUP are post-hoc transformations of extracted
features. Adult and Bank STD use logistic regression.

The head-replacement experiment trains a one-hidden-layer classifier on the frozen DFL
representation. Its measured TPR-RMS and MCDP gaps show empirical persistence, while
Theorem~\ref{thm:main-inheritance} concerns a distance-kernel RKHS class. 

\section{Additional experimental results}\label{appendix:numeric}

\subsection{Trajectories of the performance metrics during fine-tuning}

\begin{figure}[H]
\centering
\includegraphics[width=0.48\linewidth]{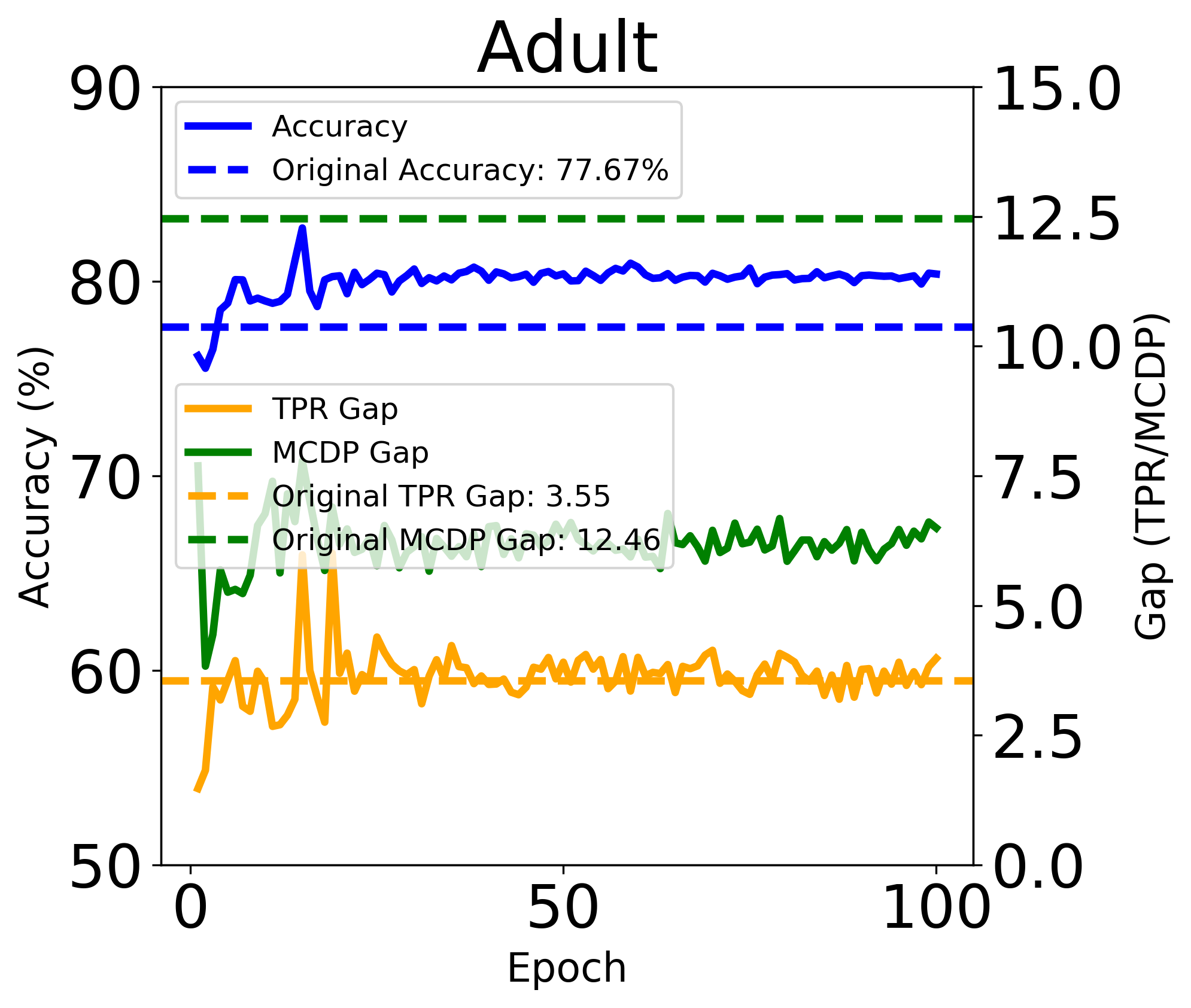}
\includegraphics[width=0.48\linewidth]{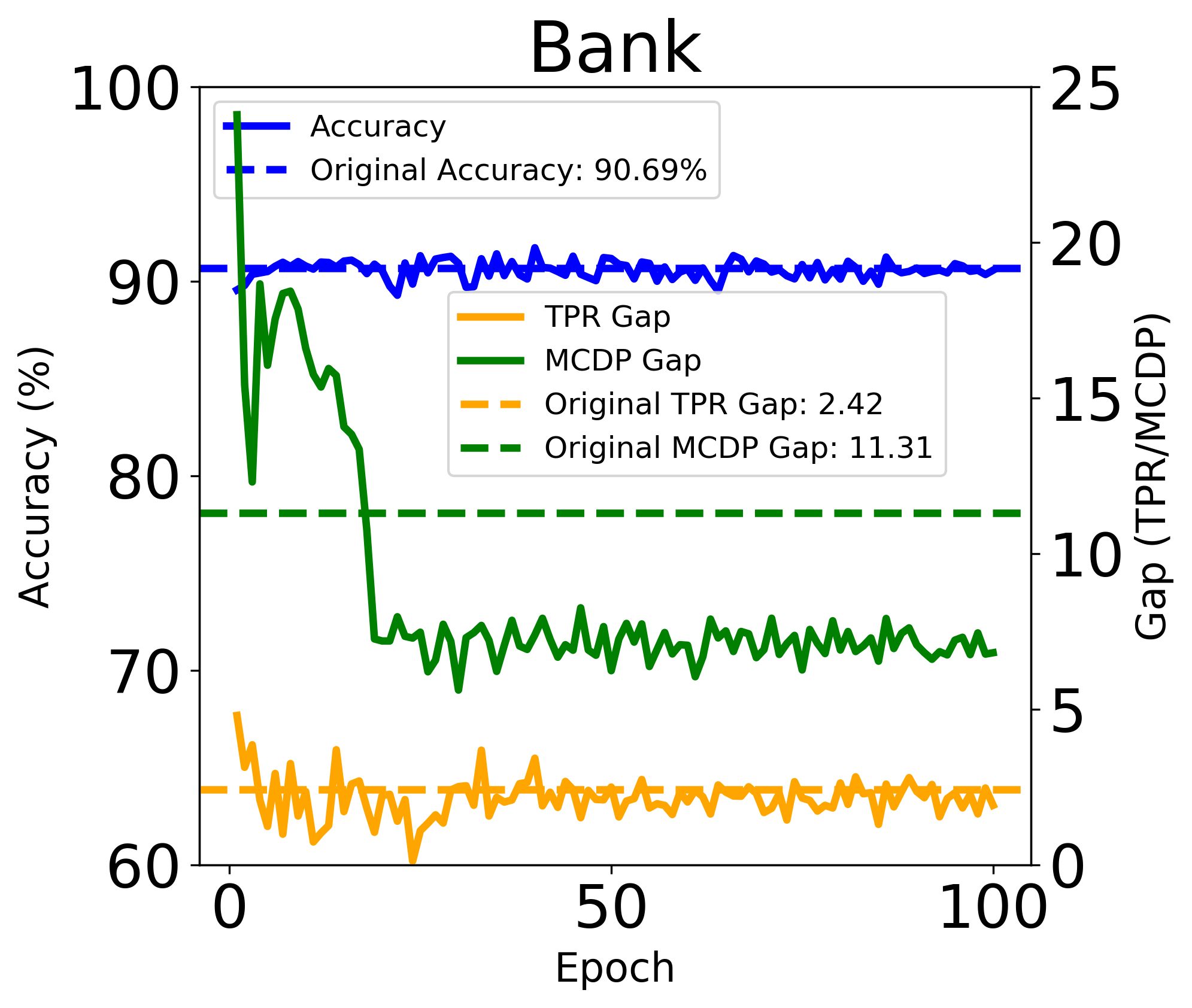}
\caption{DFL fine-tuning performance on Adult and Bank}
\label{fig:tabular}
\end{figure}

\begin{figure}[H]
\centering
\includegraphics[width=0.48\linewidth]{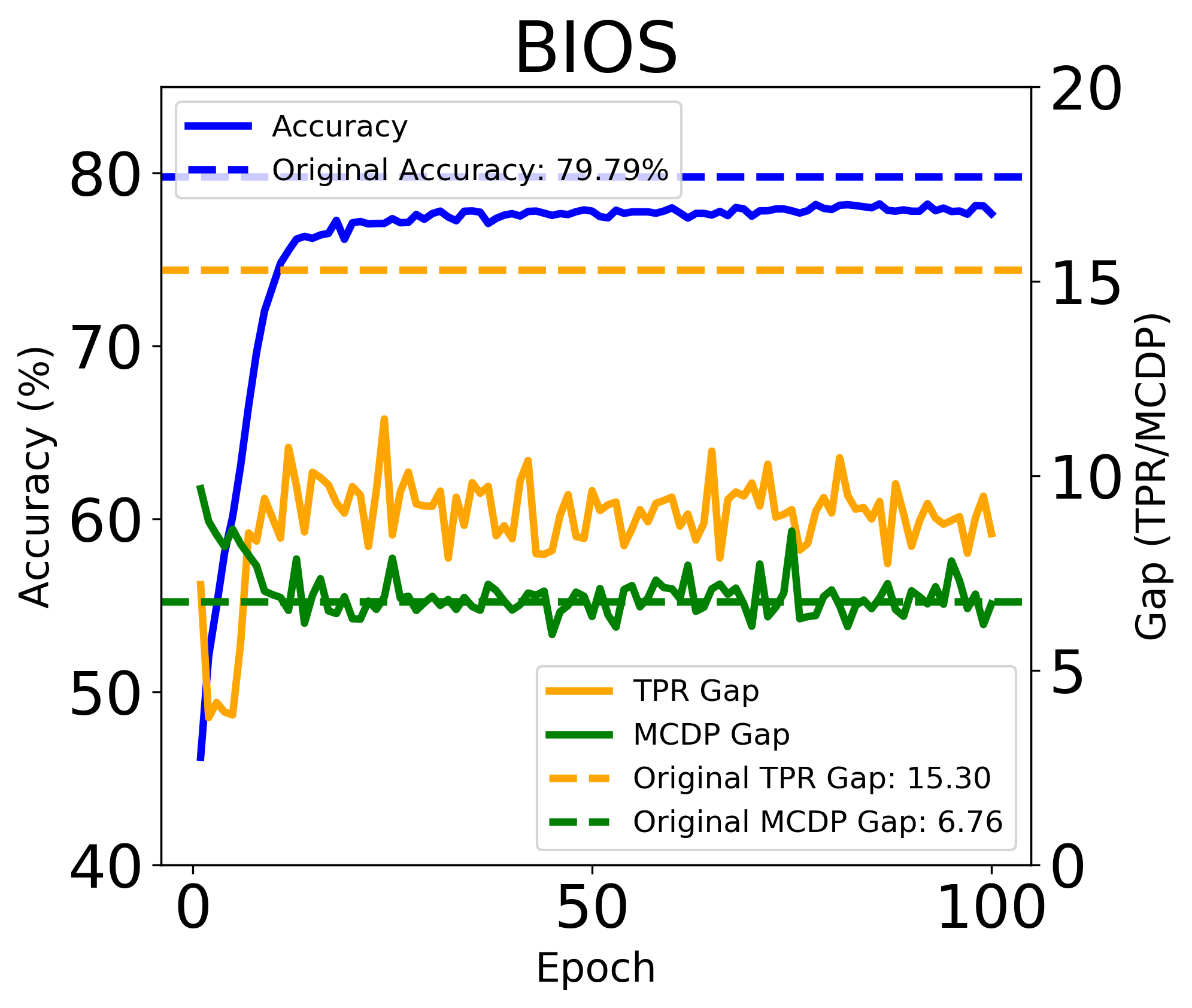}
\includegraphics[width=0.48\linewidth]{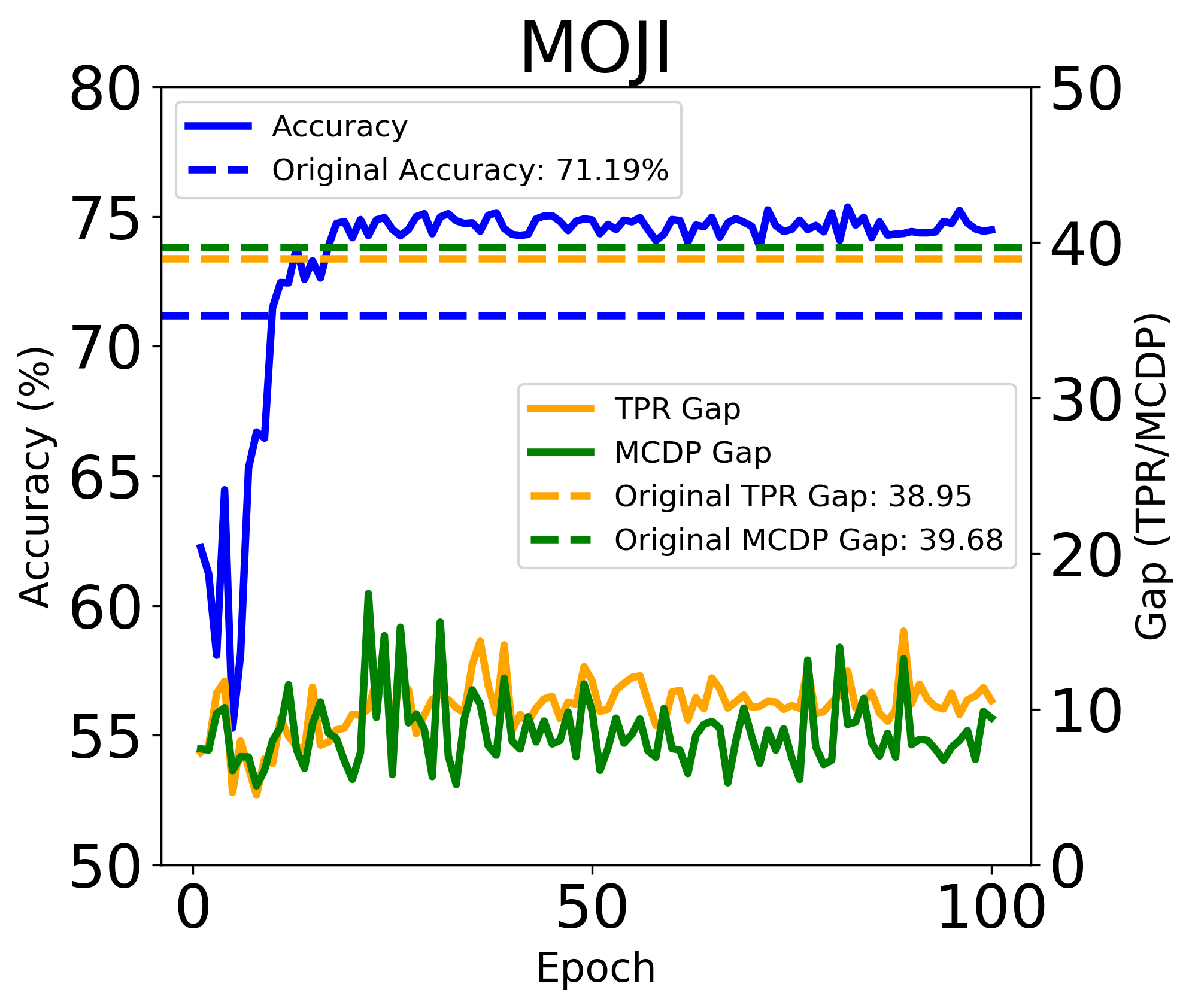}
\caption{DFL fine-tuning performance on BIOS and MOJI.}
\label{fig:nlp}
\end{figure}

We display the fine-tuning trajectories of the performance metrics (accuracy, TPR gap, and MCDP gap) for the tabular datasets (Adult and Bank) and the contextual datasets (BIOS and MOJI) in Figures~\ref{fig:tabular} and \ref{fig:nlp}, respectively.

\subsection{Additional results for CelebA}

Tables \ref{tab:celebA-blackhair} and \ref{tab:celebA-paleskin} show the results for $\bs{Z}$ = Black Hair and $\bs{Z}$ = Pale Skin across different target attributes. Tables \ref{tab:celebA-single-DFLrep} and \ref{tab:celebA-multi-DFLrep} present the results for single and
multiple sensitive attributes after discarding the DFL head and training a new unconstrained classifier on
the frozen representation.

\begin{table}[!htbp]
\centering
\caption{CelebA results with Black Hair as the sensitive attribute. Parentheses give
standard deviations. Best and second-best fairness-aware means are bold and underlined.}
\label{tab:celebA-blackhair}
\fontsize{7.5}{8.5}\selectfont
\setlength{\tabcolsep}{2.5pt}
\begin{tabular}{lc|cccccc}
\toprule
Target Attribute $\bs{Y}$ & Metric & Standard & DFL & DiffMCDP & RLACE & CFair-EO & LAFTR-EO-CE \\
\midrule
\multirow{3}{*}{Young} 
& Accuracy $\uparrow$ & 85.01 & 81.68(1.41) & 81.38(0.78) & 82.33(1.11) & \textbf{82.55(0.23)} & \underline{82.38(0.45)} \\
& TPR Gap $\downarrow$ & 19.50 & \textbf{8.39(2.47)} & 8.97(0.95) & 12.10(1.67) & 8.72(1.37) & \underline{8.55(1.12)} \\
& MCDP Gap $\downarrow$ & 11.40 & 6.18(0.83) & 8.72(0.87) & 9.07(1.31) & \underline{5.46(1.41)} & \textbf{4.73(0.98)} \\
\midrule
\multirow{3}{*}{Attractive} 
& Accuracy $\uparrow$ & 78.32 & \underline{77.01(1.09)} & 76.85(0.88) & 76.50(1.25) & 76.24(0.71) & \textbf{77.15(0.63)} \\
& TPR Gap $\downarrow$ & 5.83 & 5.78(0.43) & 5.68(0.74) & \textbf{4.62(1.44)} & \underline{5.36(0.17)} & 5.58(0.50) \\
& MCDP Gap $\downarrow$ & 5.63 & \textbf{3.13(0.66)} & 3.35(0.65) & 4.99(1.04) & 3.34(0.47) & \underline{3.19(0.46)} \\
\midrule
\multirow{3}{*}{Smiling} 
& Accuracy $\uparrow$ & 80.51 & \textbf{79.30(2.10)} & 76.91(0.74) & 78.61(1.32) & 78.82(1.13) & \underline{78.87(1.10)} \\
& TPR Gap $\downarrow$ & 0.57 & 1.06(0.71) & \textbf{0.56(0.11)} & \underline{0.65(0.98)} & 0.70(0.46) & 0.81(0.31) \\
& MCDP Gap $\downarrow$ & 1.12 & 1.98(0.49) & \textbf{1.47(0.21)} & \underline{1.60(0.92)} & 1.65(0.37) & 1.61(0.77) \\
\midrule
\multirow{3}{*}{Wavy Hair} 
& Accuracy $\uparrow$ & 83.52 & \underline{82.36(1.84)} & 80.90(0.85) & 81.55(1.06) & \textbf{82.50(0.45)} & 81.90(1.35) \\
& TPR Gap $\downarrow$ & 9.75 & \underline{1.78(0.82)} & 4.49(0.87) & 3.05(1.12) & 3.31(0.85) & \textbf{0.82(0.23)} \\
& MCDP Gap $\downarrow$ & 15.67 & \underline{7.31(0.85)} & 8.55(1.13) & 9.33(1.76) & 8.53(0.68) & \textbf{6.72(0.59)} \\
\bottomrule
\end{tabular}
\end{table}

\begin{table}[!htbp]
\centering
\caption{CelebA results with Pale Skin as the sensitive attribute. Parentheses give
standard deviations. Best and second-best fairness-aware means are bold and underlined.}
\label{tab:celebA-paleskin}
\fontsize{7.5}{8.5}\selectfont
\setlength{\tabcolsep}{2.5pt}
\begin{tabular}{lc|cccccc}
\toprule
Target Attribute $\bs{Y}$ & Metric & Standard & DFL & DiffMCDP & RLACE & CFair-EO & LAFTR-EO-CE \\
\midrule
\multirow{3}{*}{Young} 
& Accuracy $\uparrow$ & 85.01 & \textbf{84.53(0.44)} & 83.20(0.75) & 81.88(1.10) & \underline{84.23(0.50)} & 84.15(0.57) \\
& TPR Gap $\downarrow$ & 8.07 & \underline{6.62(1.32)} & 7.72(0.68) & 7.01(1.23) & \textbf{6.20(0.86)} & 6.80(0.69) \\
& MCDP Gap $\downarrow$ & 18.58 & 9.94(0.91) & 9.77(1.04) & 9.89(1.88) & \underline{8.74(1.81)} & \textbf{7.34(0.91)} \\
\midrule
\multirow{3}{*}{Attractive} 
& Accuracy $\uparrow$ & 78.32 & \underline{77.51(0.53)} & \textbf{77.96(0.81)} & 77.48(1.14) & 75.81(0.99) & 75.99(0.94) \\
& TPR Gap $\downarrow$ & 13.90 & 9.13(1.65) & 10.40(0.92) & 10.63(1.90) & \textbf{8.05(1.40)} & \underline{8.49(2.24)} \\
& MCDP Gap $\downarrow$ & 23.10 & 18.93(0.76) & \textbf{16.65(1.39)} & 18.29(2.05) & \underline{17.18(1.11)} & 19.08(1.56) \\
\midrule
\multirow{3}{*}{Smiling} 
& Accuracy $\uparrow$ & 80.51 & \textbf{79.92(1.11)} & 77.04(0.83) & \underline{79.50(0.91)} & 78.81(0.69) & 78.62(0.78) \\
& TPR Gap $\downarrow$ & 10.03 &  \underline{4.58(1.38)} & 4.61(0.91) & 6.23(1.34) & \textbf{4.37(1.10)} &4.78(1.07) \\
& MCDP Gap $\downarrow$ & 20.64 &  15.68(1.09) & \underline{15.63(1.29)} & 16.98(2.13) &15.93(1.06) & \textbf{14.89(1.24)} \\
\midrule
\multirow{3}{*}{Wavy Hair} 
& Accuracy $\uparrow$ & 83.52 & \textbf{83.20(0.50)} & \underline{82.89(0.77)} & 82.49(1.03) & 82.30(0.69) & 82.84(0.77) \\
& TPR Gap $\downarrow$ & 1.91 & \underline{1.58(0.89)} & 1.89(0.27) & \textbf{1.48(0.93)} & 1.92(0.80) & 2.51(0.95) \\
& MCDP Gap $\downarrow$ & 8.21 & 8.08(1.58) & \textbf{7.43(0.91)} & 8.37(1.43) & 8.31(0.70) & \underline{8.05(1.13)} \\
\bottomrule
\end{tabular}
\end{table}

\begin{table*}[!htbp]
\centering
\caption{Results for the CelebA dataset with a single sensitive attribute, using the frozen DFL encoder $g_{\hat\theta}$ with a freshly trained unconstrained head (the head-replacement setting of
Section~\ref{sec:rq2}).}
\label{tab:celebA-single-DFLrep}
\fontsize{8.5}{9.5}\selectfont
\setlength{\tabcolsep}{3pt}
\begin{tabular}{c@{\hskip 5pt}c|c@{\hskip 5pt}c|c@{\hskip 5pt}c|c@{\hskip 5pt}c|c@{\hskip 5pt}c}
\toprule
 &\hfill$\bs{Y}$& \multicolumn{2}{c|}{Young} & \multicolumn{2}{c|}{Attractive} & \multicolumn{2}{c|}{Smiling} & \multicolumn{2}{c}{Wavy Hair} \\
 {Metric} & $\bs{Z}$ & STD & DFL & STD & DFL & STD & DFL & STD & DFL \\
\midrule
\multirow{3}{*}{Accuracy $\uparrow$} 
& Male & 85.01 & 79.12(1.08) & 78.32 & 74.85(1.65) & 80.51 & 77.06(1.34) & 83.52 & 79.45(0.85) \\
& Black Hair & 85.01 & 81.43(0.99) & 78.32 & 76.54(1.03) & 80.51 & 78.81(1.85) & 83.52 & 81.69(1.56) \\
& Pale Skin & 85.01 & 83.98(0.67) & 78.32 & 77.15(0.89) & 80.51 & 79.35(1.42) & 83.52 & 82.91(0.68) \\
\midrule
\multirow{3}{*}{TPR Gap $\downarrow$}  
& Male &  26.70 & 4.85(2.45) &  40.50 & 5.91(1.74) &  7.65 & 6.14(2.03) &  29.14 & 5.49(2.78) \\
& Black Hair &  19.50 & 9.81(2.93) &  5.83 & 7.02(0.87) &  0.57 & 1.06(1.04) &  9.75 & 4.15(1.06) \\
& Pale Skin &  8.07 & 7.98(1.56) &  13.90 & 10.23(1.98) &  10.03 & 6.73(1.52) &  1.91 & 2.15(1.25) \\
\midrule
\multirow{3}{*}{MCDP Gap $\downarrow$} 
& Male &  49.74 & 12.98(2.85) &  57.20 & 23.54(3.12) &  16.63 & 8.94(2.04) &  48.13 & 19.54(3.02) \\
& Black Hair &  11.40 & 7.15(1.25) &  5.63 & 4.34(0.76) &  1.12 & 2.02(0.87) &  15.67 & 8.03(1.25) \\
& Pale Skin &  18.58 & 10.24(1.03) &  23.10 & 20.43(1.24) &  20.64 & 16.92(1.94) &  8.21 & 9.24(1.63) \\
\bottomrule
\end{tabular}
\end{table*}

\begin{table*}[!htbp]
\centering
\caption{Results for the CelebA dataset with multiple sensitive attributes, using the frozen DFL encoder $g_{\hat\theta}$ with a freshly trained unconstrained head.}
\label{tab:celebA-multi-DFLrep}
\fontsize{8.5}{9.5}\selectfont
\setlength{\tabcolsep}{3pt}
\begin{tabular}{c@{\hskip 5pt}c|c@{\hskip 5pt}c|c@{\hskip 5pt}c|c@{\hskip 5pt}c|c@{\hskip 5pt}c}
\toprule
 &\hfill$\bs{Y}$& \multicolumn{2}{c|}{Young} & \multicolumn{2}{c|}{Attractive} & \multicolumn{2}{c|}{Smiling} & \multicolumn{2}{c}{Wavy Hair} \\
 {Metric} & $\bs{Z}$ & STD & DFL & STD & DFL & STD & DFL & STD & DFL \\
\midrule
\multirow{3}{*}{Accuracy $\uparrow$} 
& Male & 85.01 & 81.36(1.12) & 78.32 & 75.14(1.45) & 80.51 & 79.54(1.02) & 83.52 & 82.11(1.06) \\
& Black Hair & 85.01 & 81.36(1.12) & 78.32 & 75.14(1.45) & 80.51 & 79.54(1.02) & 83.52 & 82.11(1.06) \\
& Pale Skin & 85.01 & 81.36(1.12) & 78.32 & 75.14(1.45) & 80.51 & 79.54(1.02) & 83.52 & 82.11(1.06) \\
\midrule
\multirow{3}{*}{TPR Gap $\downarrow$}  
& Male & 26.73 & 5.94(3.74) & 40.50 & 15.12(3.11) & 7.64 & 7.35(2.72) & 29.13 & 13.24(2.85) \\
& Black Hair & 19.52 & 15.02(1.87) & 5.84 & 5.94(1.64) & 0.57 & 1.15(1.43) & 9.75 & 5.31(1.84) \\
& Pale Skin & 8.01 & 3.27(1.82) & 13.90 & 11.35(1.88) & 10.02 & 9.32(0.83) & 1.90 & 2.81(0.85) \\
\midrule
\multirow{3}{*}{MCDP Gap $\downarrow$} 
& Male & 49.73 & 24.14(5.21) & 57.20 & 34.62(4.83) & 16.64 & 14.01(5.83) & 48.12 & 30.47(4.95) \\
& Black Hair & 11.40 & 11.56(2.45) & 5.63 & 6.14(1.88) & 1.14 & 2.51(1.41) & 15.60 & 10.52(1.79) \\
& Pale Skin & 18.52 & 14.11(1.84) & 23.10 & 20.34(1.95) & 20.64 & 19.81(2.62) & 8.21 & 6.41(1.92) \\
\bottomrule
\end{tabular}
\end{table*}

\subsection{Multiple sensitive attributes on CelebA}
\label{app:multi-celeba}

Table~\ref{tab:celebA-multi} reports the multi-attribute setting of Section~\ref{sec:rq5}, in which
$\bs{Z}$ concatenates Male, Black Hair and Pale Skin and the conditional DC penalty is applied to the
three-dimensional vector. Gaps are computed per coordinate, with the other coordinates marginalized out.
These are marginal and not intersectional quantities: they do not measure disparity
on the eight subgroups defined by the attribute combinations, and the reductions below should not be read as
evidence about those subgroups. On most coordinates the reduction is smaller than in the corresponding single-attribute run, which is
expected when one penalty must serve three attributes; two Pale Skin cells go the other way (Young TPR falls
$8.01\to1.97$ jointly against $8.07\to6.62$ singly, and Wavy Hair MCDP falls $8.21\to5.41$ against
$8.21\to8.08$), and on a few already-small cells the gap increases slightly.

\begin{table*}[!htbp]
\centering
\caption{Results for the CelebA dataset with multiple sensitive attributes using DFL. Mean (standard deviation) over 20
replications. Marginal per-coordinate gaps; see Appendix~\ref{app:multi-metrics} for why these are not
intersectional.}
\label{tab:celebA-multi}
\fontsize{8.5}{9.5}\selectfont
\setlength{\tabcolsep}{3pt}
\begin{tabular}{c@{\hskip 2pt}c|c@{\hskip 5pt}c|c@{\hskip 5pt}c|c@{\hskip 5pt}c|c@{\hskip 5pt}c}
\toprule
 &\hfill$\bs{Y}$& \multicolumn{2}{c|}{Young} & \multicolumn{2}{c|}{Attractive} & \multicolumn{2}{c|}{Smiling} & \multicolumn{2}{c}{Wavy Hair} \\
 {Metric} & $\bs{Z}$ & STD & DFL & STD & DFL & STD & DFL & STD & DFL \\
\midrule
\multirow{3}{*}{Accuracy $\uparrow$}
& Male & 85.01 & 81.79(0.83) & 78.32 & 75.87(0.90) & 80.51 & 80.31(0.66) & 83.52 & 82.70(0.46) \\
& Black Hair & 85.01 & 81.79(0.83) & 78.32 & 75.87(0.90) & 80.51 & 80.31(0.66) & 83.52 & 82.70(0.46) \\
& Pale Skin & 85.01 & 81.79(0.83) & 78.32 & 75.87(0.90) & 80.51 & 80.31(0.66) & 83.52 & 82.70(0.46) \\
\midrule
\multirow{3}{*}{TPR Gap $\downarrow$}
& Male & 26.73 & 3.94(3.41) & 40.50 & 13.45(4.79) & 7.64 & 6.93(2.12) & 29.13 & 12.08(4.39) \\
& Black Hair & 19.52 & 12.89(1.67) & 5.84 & 6.03(1.09) & 0.57 & 1.65(1.19) & 9.75 & 4.68(1.63) \\
& Pale Skin & 8.01 & 1.97(1.54) & 13.90 & 10.11(1.29) & 10.02 & 8.42(2.58) & 1.90 & 1.92(0.68) \\
\midrule
\multirow{3}{*}{MCDP Gap $\downarrow$}
& Male & 49.73 & 23.31(3.69) & 57.20 & 32.58(3.58) & 16.64 & 13.36(2.63) & 48.12 & 29.36(3.29) \\
& Black Hair & 11.40 & 10.03(1.92) & 5.63 & 5.24(1.29) & 1.14 & 2.65(1.19) & 15.60 & 9.12(1.58) \\
& Pale Skin & 18.52 & 13.04(1.51) & 23.10 & 19.02(1.47) & 20.64 & 19.38(2.21) & 8.21 & 5.41(1.68) \\
\bottomrule
\end{tabular}
\end{table*}

\subsection{Ablation studies}\label{sec:ablation-studies}

We examine the contribution of the DFL objective components on BIOS using the following terms:
\begin{itemize}
    \item $T_1 = \mathbb{E}[\ell(\bs{Y}, f_{\phi}(g_{\theta}(\bs{X})))]$,
    \item $T_2 = \operatorname{DC}(\mathbf{e}_{\bs{Y}}, g_{\theta}(\bs{X}))$,
    \item $T_3 = \operatorname{DC}(\bs{Z}, g_{\theta}(\bs{X}) \mid \bs{Y})$,
    \item $T_4 = \operatorname{DC}(\bs{Z}, f_{\phi}(g_{\theta}(\bs{X})) \mid \bs{Y})$.
\end{itemize}
We evaluate six loss functions:
\begin{align*}
    L_1 &= \alpha(T_1 - T_2) + (1 - \alpha) T_3 \quad \text{(Original DFL)} \\
    L_2 &= \alpha T_1 + (1 - \alpha) T_3 \quad \text{(Without $T_2$)} \\
    L_3 &= \alpha(T_1 - T_2) + (1 - \alpha)(T_3 + T_4) \quad \text{(Adding $T_4$)} \\
    L_4 &= \alpha T_1 + (1 - \alpha)(T_3 + T_4) \quad \text{(Without $T_2$, with $T_4$)} \\
    L_5 &= \alpha T_1 + (1 - \alpha) T_4 \quad \text{(Without $T_2, T_3$, only $T_4$)}\\
    L_6 &= -\alpha T_2 + (1 - \alpha) T_3 \quad \text{(Without downstream loss)}
\end{align*}

The results, summarized in Table~\ref{tab:ablation}, report accuracy, TPR gap, and MCDP gap averaged over five replications.

\begin{table}[t]
\centering
\caption{Effect of the objective components on BIOS. Mean (standard deviation) over 5 replications, all other settings fixed.
$L_2$ versus $L_5$ isolates representation-level against prediction-level dependence control; $L_1$ versus
$L_2$ isolates the informativeness term $T_2$.}
\label{tab:ablation}
\fontsize{8}{9.5}\selectfont
\setlength{\tabcolsep}{4pt}
\begin{tabular}{@{}llccc@{}}
\toprule
\textbf{Objective} & \textbf{Composition} & \textbf{Accuracy $\uparrow$} & \textbf{TPR gap $\downarrow$} & \textbf{MCDP gap $\downarrow$} \\
\midrule
$L_1$ & $\alpha(T_1-T_2)+(1-\alpha)T_3$ (full objective) & 78.09 (0.42) & 9.08 (0.65) & 6.21 (0.81) \\
$L_2$ & $\alpha T_1+(1-\alpha)T_3$ (no informativeness term) & 77.85 (0.38) & 8.97 (0.74) & 6.52 (0.71) \\
$L_3$ & $\alpha(T_1-T_2)+(1-\alpha)(T_3+T_4)$ & 61.46 (0.33) & 7.35 (0.69) & 6.99 (0.66) \\
$L_4$ & $\alpha T_1+(1-\alpha)(T_3+T_4)$ & 60.52 (0.49) & 4.28 (0.75) & 8.52 (0.77) \\
$L_5$ & $\alpha T_1+(1-\alpha)T_4$ (prediction-level DC only) & 64.78 (0.36) & 5.12 (0.70) & 7.76 (0.62) \\
$L_6$ & $-\alpha T_2+(1-\alpha)T_3$ (no task loss) & 74.79 (0.51) & 11.01 (0.62) & 7.83 (0.74) \\
\bottomrule
\end{tabular}
\end{table}

The primary comparisons are penalty location ($L_2$ versus $L_5$) and the informativeness term
($L_1$ versus $L_2$), discussed in Sections~\ref{sec:rq3} and~\ref{sec:rq4}. These are comparisons at
one operating point rather than complete fairness--utility frontiers. The $L_6$ objective contains no
head parameters, and the head-fitting procedure for its reported accuracy is undocumented. That row
cannot establish the performance of a controlled two-stage procedure.
Removing $T_2$ changes accuracy from $78.09\%$ to $77.85\%$ at this setting, slightly reduces
the TPR-RMS gap, and raises MCDP. This mixed effect does not establish that $T_2$ is essential
at other operating points.

\subsection{Hyperparameter sensitivity}\label{app:sensitivity}

\begin{table}[t]
\centering
\caption{Sensitivity to the fairness--utility weight on BIOS. Mean (standard deviation) over 5 replications; depth 10, growth rate 64 throughout.}
\label{tab:alpha}
\fontsize{8.5}{9.5}\selectfont
\begin{tabular}{cccc}
\toprule
$\alpha$ & \textbf{Accuracy $\uparrow$} & \textbf{TPR gap $\downarrow$} & \textbf{MCDP gap $\downarrow$} \\
\midrule
0.1 & 48.53 (0.39) & 4.77 (0.51) & 5.38 (0.63) \\
0.3 & 62.33 (0.48) & 8.76 (0.66) & 6.31 (0.61) \\
0.5 & 71.67 (0.37) & 8.52 (0.72) & 6.04 (0.68) \\
0.7 & 77.82 (0.52) & 9.44 (0.62) & 6.42 (0.69) \\
0.9 & 78.76 (0.47) & 12.02 (0.73) & 6.83 (0.71) \\
\bottomrule
\end{tabular}
\end{table}

\begin{remark}[Cross-table variation in separate sweeps]
\label{rem:inconsistencies}
Two differences affect cross-table comparisons. First, the configuration
$\alpha=0.7$, depth $10$, growth rate $64$ appears both in
Table~\ref{tab:alpha} ($77.82/9.44/6.42$) and in Table~\ref{tab:hyperparam-performance}
below ($77.67/9.52/6.04$). These were separate sweeps, and the differences lie within
the reported replication standard deviations; the entries are independent summaries
rather than repeated measurements from an identical run.
Second, the STD baseline for a given $(\bs Y,\bs Z)$ cell differs by up to $0.07$ between tables (for
example, Male/Young TPR is $26.70$ in Table~\ref{tab:celebA-male} and $26.73$ in
Table~\ref{tab:celebA-multi}). These tables retain their separately evaluated STD references;
small cross-table differences should not be interpreted as treatment effects.
\end{remark}

To assess the robustness of DFL, we perform a sensitivity analysis with respect to two key aspects: the loss trade-off parameter \(\alpha\) and the architecture of the DenseNet backbone.

\textbf{Effect of Trade-off Parameter \(\alpha\)}

The parameter \(\alpha \in [0, 1]\) controls the balance between the utility term (cross-entropy loss) and the fairness constraint in the DFL objective. Table~\ref{tab:alpha} reports the performance of DFL on the BIOS dataset across various values of \(\alpha\), with all other settings held fixed.

As \(\alpha\) increases, more weight is placed on the prediction loss. This leads to improved accuracy, from 48.53\% at \(\alpha = 0.1\) to 78.76\% at \(\alpha = 0.9\), but comes at the cost of fairness. Specifically, both the TPR gap and the MCDP gap increase overall, though not monotonically --- both dip at $\alpha=0.5$
--- reflecting reduced fairness enforcement. This illustrates the expected fairness–utility trade-off and highlights \(\alpha = 0.7\) as a balanced choice, offering competitive accuracy (77.82\%) while keeping fairness metrics at acceptable levels.

\textbf{Effect of Network Architecture}

We further analyze the impact of the DenseNet encoder’s architecture by varying its depth and growth rate, summarized in Table~\ref{tab:hyperparam-performance}. All experiments are conducted with \(\alpha = 0.7\), the reference setting used in this sweep.

Increasing the growth rate (i.e., the width of the network) improves overall performance, with the best fairness observed at a growth rate of 96. For example, increasing the growth rate from 32 to 96 (at depth 10) reduces the MCDP gap from 6.51 to 5.53 and marginally improves accuracy. Increasing depth helps accuracy but not fairness uniformly: at growth rate $64$ the TPR gap rises from
$9.52$ at depth $10$ to $10.21$ at depth $20$. The combination of depth $20$ and growth rate $96$ attains the
highest accuracy ($78.86\%$) with an MCDP gap of $5.63$.

These results indicate that DFL's behaviour is not an artifact of one network size: over a $7.5\times$ range
of parameter counts accuracy moves by $1.4$ points and the MCDP gap by $1.0$. Larger encoders help slightly,
but not uniformly --- at growth rate $64$, going from depth $10$ to depth $20$ raises the TPR gap from $9.52$
to $10.21$. 

\begin{table}[!htbp]
\centering
\caption{Performance of DFL on BIOS dataset under different DenseNet architecture (growth rates and depths). Values are reported as mean (std) for 5 replications.}
\label{tab:hyperparam-performance}
 \fontsize{8.5}{9.5}\selectfont
\begin{tabular}{ccccccc}
\toprule
\textbf{Alpha} & \textbf{Depth} & \textbf{Growth Rate} & \textbf{Parameters} & \textbf{Acc $\uparrow$} & \textbf{TPR Gap $\downarrow$} & \textbf{MCDP Gap $\downarrow$} \\
\midrule
0.7 & 10 & 32 & 132139  & 77.45 (0.44) & 9.93 (0.64) & 6.51 (0.76) \\
0.7 & 10 & 64 & 295030  & 77.67 (0.37) & 9.52 (0.72) & 6.04 (0.68) \\
0.7 & 10 & 96 & 510973  & 77.82 (0.33) & 9.49 (0.71) & 5.53 (0.70) \\
0.7 & 20 & 32 & 206187  & 78.42 (0.36) & 9.41 (0.67) & 5.84 (0.63) \\
0.7 & 20 & 64 & 529775  & 78.61 (0.41) & 10.21 (0.75) & 6.11 (0.53) \\
0.7 & 20 & 96 & 991071  & 78.86 (0.39) & 10.03 (0.69) & 5.63 (0.66) \\
\bottomrule
\end{tabular}
\end{table}

\end{document}